%% file: neurips_2026.tex
\documentclass{article}

\usepackage[final,main]{neurips_2026}

\usepackage[utf8]{inputenc} % allow utf-8 input
\usepackage[T1]{fontenc}    % use 8-bit T1 fonts
\usepackage{hyperref}       % hyperlinks
\usepackage{url}            % simple URL typesetting
\usepackage{booktabs}       % professional-quality tables
\usepackage{amsmath}        % \tfrac, equation environments
\usepackage{amsfonts}       % blackboard math symbols
\usepackage{nicefrac}       % compact symbols for 1/2, etc.
\usepackage{microtype}      % microtypography
\usepackage{xcolor}         % colors
\usepackage{graphicx}       % figures
\usepackage{wrapfig}        % wrapped role figure and ablations table (\S3.3--3.4)
\usepackage{needspace}      % keep headings with the text that follows
\usepackage{subcaption}     % side-by-side subfigures (LawBench example in \S2)
\usepackage{algorithm}      % evolutionary orchestrator pseudocode (appendix)
\usepackage{algpseudocode}
\usepackage{comment}        % \begin{comment} ... \end{comment} to fold out template prose

\definecolor{LightBlue}{RGB}{220,230,250}
\definecolor{LightRed}{RGB}{250,220,220}

\definecolor{Red}{rgb}{0.768, 0.054, 0.054}
\definecolor{Blue}{rgb}{0.152, 0.294, 0.925}
\definecolor{Green}{rgb}{0,0.4,0.7}

\hypersetup{
    colorlinks=true,
    citecolor=teal,
    linkcolor=Red,
    urlcolor=Green,
}

\newcommand{\metan}{\texorpdfstring{Meta\textsuperscript{\textit{n}}}{Meta-n}}

\title{\metan: Recursive Self-Improvement \\ through Emergent Depth}

\author{
  Zae Myung Kim\textsuperscript{1} \quad
  Young-Jun Lee\textsuperscript{1} \quad
  Seungyeon Jwa\textsuperscript{2} \quad
  Dongyeop Kang\textsuperscript{1} \\[0.4em]
  \textsuperscript{1}University of Minnesota \quad
  \textsuperscript{2}Seoul National University \\[0.2em]
  \texttt{\{kim01756, lee05727, dongyeop\}@umn.edu} \quad
  \texttt{amyj97@snu.ac.kr}
}

\begin{document}

\maketitle

\input{sections/0_abstract}

\input{sections/1_introduction}

\input{sections/2_method}

\input{sections/3_experiments}

\input{sections/4_conclusion}

\section*{Acknowledgment}
We thank members of Minnesota NLP for their insightful comments during group meetings. ZMK is generously supported by the 3M Science and Technology Fellowship and the Doctoral Dissertation Fellowship at the University of Minnesota.

\bibliographystyle{plainnat}
\bibliography{references}

\appendix
\newpage
\input{sections/9_appendix}

\clearpage
\input{checklist}

\end{document}

%% file: sections/0_abstract.tex
\begin{abstract}
Self-improving LLM agents refine answers, not the \emph{process} that produces those answers. Systems that add a meta-level hold that level fixed, and those that edit themselves must leave part of their own editing machinery untouched to stay stable, capping the \emph{meta-depth} they realize at roughly two. We present \textbf{\metan{}}, which keeps the meta-operation fixed and recurses on its input instead. That operation, $\Omega$, is applied repeatedly to its own products, reading the traces of the solver stack below together with the code that produced them, then writing the next layer as a strategic pre-process and a library of callable helpers. Because $\Omega$ never changes, it cannot destabilize the system, and because its input strictly grows, each layer reasons from a higher vantage than the last. Depth is set by convergence rather than fixed in advance, and an evolutionary archive searches over layer chains. Across two backbones, \metan{} outperforms prior self-improving agents on all eight benchmark families. The sharpest case is ARC-AGI-2, built to resist skill memorization, where it alone scores above zero. Ablations indicate that most of the gain from recursion comes from the conditioning each layer passes to the next, and distinct layer roles emerge with depth although no prompt prescribes them.
Code available at \url{https://github.com/minnesotanlp/meta-n}
\end{abstract}

%% file: sections/1_introduction.tex
\begin{figure}[h]
    \vspace{-3mm}
    \centering
    \includegraphics[width=0.95\linewidth]{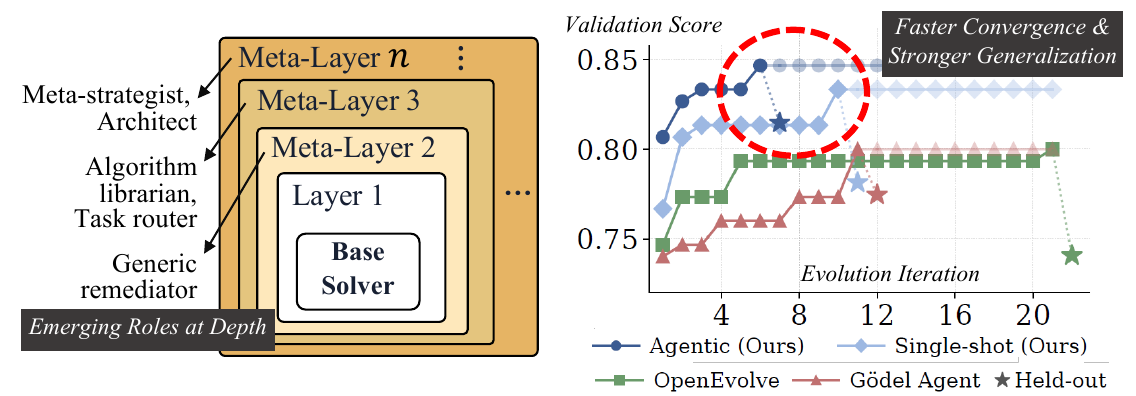}
    \caption{\textbf{\metan{} at a glance.} \emph{(left)}~A single universal meta-operation $\Omega$ is applied recursively, each meta-layer wrapping the stack below it (Base Solver $\to$ Layer~1 $\to$ Meta-Layers $2,3,\dots,n$). Roles emerge with depth without manual design. \emph{(right)}~Search progress on LawBench charge prediction, where both \metan{} variants climb faster and settle higher than G\"{o}del Agent \citep{godelAgent2024} and OpenEvolve \citep{openevolve}.}
    \label{fig:hook}
\end{figure}

%==============================================================================
\section{Introduction}
\label{sec:intro}
%==============================================================================

A key component of intelligence is the capacity for meta-reasoning, the ability to step outside one's own chain of reasoning and arrive at insights that lie beyond it~\citep{russell1991principles, kounios2014insight, ackerman2017metareasoning}. But how can a system get outside itself? In \emph{G\"{o}del, Escher, Bach}~\citep{hofstadter1979geb}, Hofstadter argues that intelligence and the sense of self emerge not from any single rule but from \emph{strange loops}, that is, self-referential processes in which a system's own rules, applied recursively, fold back to operate on their own products. Today's self-improving LLM agents fall short of exactly this. They write algorithms for NP-hard optimization, operate terminal environments, and propose mathematical constructions, yet the dominant response to failure is self-refinement~\citep{selfrefine2023, reflexion2023} (examine, revise, retry), a single layer of reflection in which the same mechanism generates, diagnoses, and repairs. \textbf{Existing agents improve answers, not the process by which answers are improved.}

% Prior self-improvement work falls into three families, mapped onto Figure~\ref{fig:paradigms} and Table~\ref{tab:comparison} (full discussion in Appendix~\ref{app:related_work_full}).
Recent work adds an explicit \textit{meta}-level: evolutionary methods maintain populations of candidate programs~\citep{funsearch2024, alphaevolve2025, openevolve}; self-modifying agents rewrite portions of their own source code~\citep{godelAgent2024, darwinGodel2025, hyperagents2026, stop2024}; meta-scaffolding systems search over the infrastructure around a base model~\citep{adas2025, aflow2025, metaharness2026, promptbreeder2024, agrawal2026gepa, liu2026evox}. \textbf{But the meta-mechanism itself is fixed.} Search loops, mutation operators, and selection rules are not revised, and every system that does edit itself must hold some driver layer fixed for stability, capping realized meta-depth at ${\sim}2.5$. We make these terms precise in §\ref{sec:method:paradigms} (Figure~\ref{fig:paradigms}; full comparison in Appendix~\ref{app:related_work_full}).

We build \textbf{\metan{}} around this principle. Rather than rewriting the improver itself, we hold a single universal meta-operation $\Omega$ fixed and apply it again and again to its own outputs, \textit{``given the full execution context of the system below you, write code that improves it.''} The loop folds back through $\Omega$'s own accumulated products. Because $\Omega$ takes its own prior outputs as input, each level sees more of the system beneath it than the last, moving from surface bugs to strategic choices to meta-strategic reasoning~(Figure~\ref{fig:hook}). As an isolation check, removing recursion from \metan{} drops CO-Bench archive-best validation from 0.845 to 0.714, a $+0.131$ gain from recursion alone. The effect reproduces on a second backbone and a second benchmark, and is largest where the inner loop leaves the most room (\S\ref{sec:experiments}).
\vspace{-3mm}
\paragraph{A concrete example.}
On CO-Bench's \texttt{assignment\_problem}, a single LLM call produces a script that imports \texttt{scipy.optimize}; the sandbox lacks scipy, the script crashes, the task scores 0.0. A flat self-refinement loop retries variations of the same import-based solution. \metan{}'s depth-2 $\Omega$, looking at all 36 tasks at once, observes that \emph{several} fail with similar dependency errors and decides to emit a pre-process \emph{plus} a code-library helper: ``do not use scipy; here is a \texttt{validate\_output()} utility.'' The depth-2 solver re-implements the Hungarian algorithm and scores 1.0. Depth 3 then sees both the new traces and the depth-2 code that produced them, and determines whether to keep, refine, or override. A flat loop, seeing only traces, cannot perform such reasoning.
Repeated applications of $\Omega$ build a stack in which each layer is both a \emph{producer} of code (guidance and library functions) and the \emph{subject} of the next $\Omega$ call. We do not pre-assign layer depth, but let it grow until $\Omega$ stops finding improvements. Deeper layers learn \textbf{when} to apply, suppress, or combine primitives, because each $\Omega$ call sees the prior layers' code together with how it performed.

Across eight benchmarks spanning combinatorial optimization, text classification (Symptom2Disease and LawBench), terminal agent tasks, mathematical discovery (AlphaEvolve Math and ARC-AGI-2), symbolic regression, and algorithm speedup, \metan{} substantially improves single-call solvers without within-task iteration. Its largest margins over prior self-improving agents come on the hardest benchmarks. On ARC-AGI-2~\citep{chollet2026arcagi2newchallengefrontier}'s held-out split, for instance, \metan{} is the only system to solve any task at all, where both OpenEvolve and G\"{o}del Agent solve none.

\textbf{Our contributions.}
(1)~\textbf{The \metan{} framework (\S\ref{sec:method}):} a recursive architecture in which one universal meta-operation $\Omega$, applied repeatedly, builds a hierarchical agent stack whose depth is set by convergence. To our knowledge, the first demonstration that meta-depth beyond two yields structurally distinct levels rather than redundant ones.
(2)~\textbf{Evolutionary orchestration (\S\ref{sec:method:orchestration}):} a multi-candidate archive search over layer chains.
(3)~\textbf{Empirical evidence (\S\ref{sec:experiments}):} across eight benchmarks and two backbones, \metan{} outperforms prior self-improving agents on every benchmark family, with the largest margins on the hardest held-out tasks.

%% file: sections/2_method.tex
%==============================================================================
\section{\metan{}: Evolutionary Meta-Recursion}
\label{sec:method}
%==============================================================================

\paragraph{Overview.}
We build \metan{} from a single fixed meta-operation $\Omega$, applied recursively to its own previous outputs. Each application inspects how the solver stack below performed across the whole task set, then writes the next layer's code, a short Python \emph{pre-process} that injects strategic context before each task plus a small \emph{library of reusable helper functions} the solver may call. The next application sees both the new performance traces and \emph{the code that produced them}, so it can judge whether the previous layer's idea was a good one, rather than only whether an answer was wrong. The stack deepens by itself and halts when progress does. §\ref{sec:method:paradigms} organizes existing meta-improvement approaches into paradigms and locates this design as a new one among them, §\ref{sec:metalayer} defines a single \metan{} layer, §\ref{sec:method:recursion} analyzes what recursion gains over the single layer, and §\ref{sec:method:orchestration} presents the orchestrators that grow the stack.

\vspace{-3mm}
\paragraph{Problem formulation.}
A benchmark supplies $N$ tasks $\mathcal{T} = \{t_1, \dots, t_N\}$ and an evaluator that scores any candidate script on a task, $\mathrm{eval}(t_i, s) \in [0,1]$ (on LawBench, one such task is \emph{charge prediction}, scored by F1). Every solver run leaves an \emph{execution trace} $\tau$ (the script that ran, its stdout/stderr, exit code, score, and any evaluator feedback) that $\Omega$ can later inspect to diagnose what happened. A \textbf{base solver} $S_1$ is any one-shot or agentic LLM procedure that maps a task $t_i$ to its trace $\tau_i^{(1)}$; in our experiments $S_1$ is either a single LLM call or an 8-turn observe/act loop, but \metan{} is agnostic to this choice. Our goal is a stack of solvers $\{S_d\}_{d=2}^{n}$, each one wrapping the one below, that maximizes the mean score $\bar S = \tfrac{1}{N}\sum_i \mathrm{score}(\tau_i^{(d)})$. The depth $n$ is not fixed in advance; it grows until improvements stop (symbol glossary in Appendix Table~\ref{tab:glossary}).

%------------------------------------------------------------------------------
\subsection{Paradigms of Meta-Improvement}
\label{sec:method:paradigms}
%------------------------------------------------------------------------------

\begin{figure}[t]
    \centering
    \includegraphics[width=0.85\linewidth]{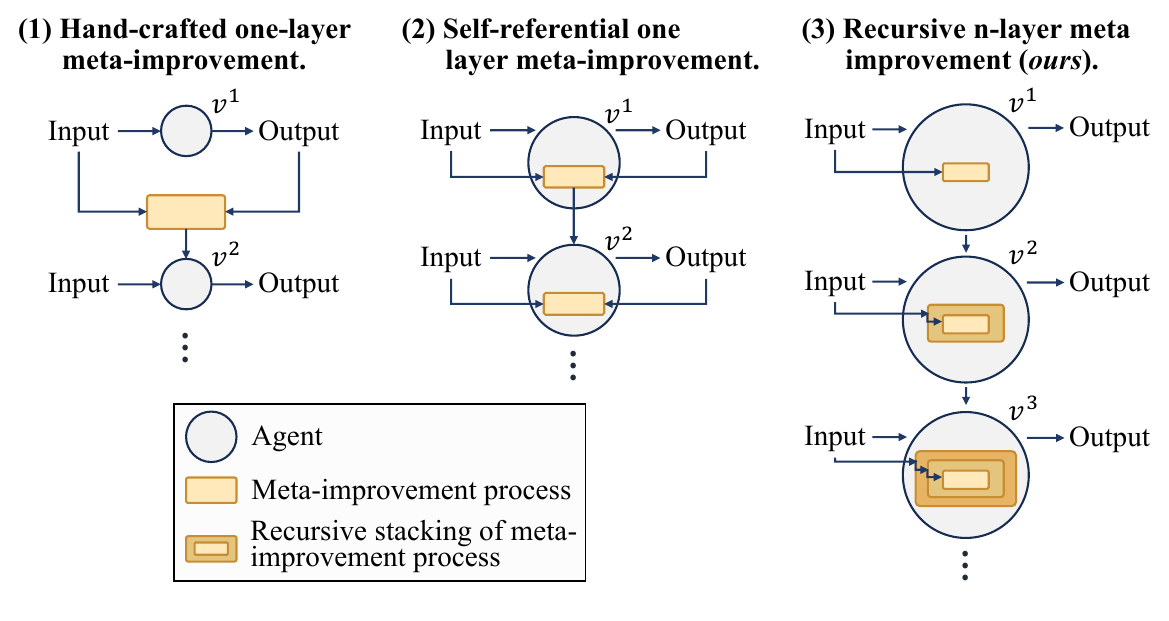}
    \caption{\textbf{Three paradigms of meta-improvement.} Circles are agents (successive versions $v^1, v^2, \dots$); amber boxes are meta-improvement processes, and \emph{nested} boxes denote recursively stacked meta-processes. \textbf{(1)~Hand-crafted one-layer meta} (e.g., FunSearch, AlphaEvolve, ADAS): a fixed, external meta-process rewrites the agent, but the meta-mechanism itself never changes. \textbf{(2)~Self-referential one-layer meta} (e.g., G\"{o}del Agent, DGM): the agent edits its own source, yet a driver layer stays fixed, capping realized meta-depth at ${\sim}2.5$. \textbf{(3)~Recursive $n$-layer meta} (\metan{}, ours): one meta-operation $\Omega$ is applied at every depth and \emph{accumulates}, so version $v^d$ carries $d$ nested meta-processes; depth is set by convergence, and each added layer observes a broader landscape of trajectories than the one below.}
    \label{fig:paradigms}
    \vspace{-3mm}
\end{figure}

To make ``depth of self-improvement'' comparable across systems, we fix three terms. Relative to a level-0 \emph{solver}, a level-1 process modifies the solver, a level-2 process modifies the level-1 process, and so on. A \emph{driver} is a component that controls such modification but itself never modified during a run. The \emph{realized meta-depth} of a system is then the highest level whose behavior actually changes over the course of a run, as opposed to the depth its architecture nominally permits. Under these terms, prior self-improving systems fall into two paradigms, and \metan{} constitutes a third (Figure~\ref{fig:paradigms}).

\textbf{(1) Hand-crafted meta-systems} place a fixed search or evolution loop \emph{outside} the solver. Evolutionary program search maintains populations of candidate programs~\citep{funsearch2024, alphaevolve2025, openevolve}; meta-scaffolding systems search over the prompts, workflows, or agent designs around a base model~\citep{adas2025, aflow2025, metaharness2026, promptbreeder2024, agrawal2026gepa}. These systems are stable and auditable precisely because the meta-process is frozen; search loops, mutation operators, and selection rules are never revised. The solver improves; the improver does not. Realized meta-depth is 1.

\textbf{(2) Self-referential agents} internalize the meta-process by letting the agent edit its own source code~\citep{godelAgent2024, darwinGodel2025, stop2024, hyperagents2026}. In principle this permits unbounded depth; in practice every such system holds some driver fixed to keep from corrupting itself: G\"{o}del Agent's action API, DGM's archive maintenance and parent selection, HyperAgents' outer evaluation loop. The editable surface is therefore a strict subset of the improvement machinery, capping realized meta-depth at ${\sim}2.5$. The agent modifies the solver (level 1), and because the modification logic lives inside the source being edited, self-edits also change how future self-edits proceed (level 2). But as the frozen driver sits inside every higher level, nothing above level 2 ever changes in full; thus the half-credit (i.e., 0.5) records this partial modifiability (per-system details are provided in Appendix~\ref{app:related_work_full}).

Together the two paradigms expose a dilemma. \emph{Recursing the improver buys depth only at the price of stability}, and (to the best of our knowledge,) every extant system resolves the tension by freezing a driver, which caps the very depth the recursion was meant to deliver.

\textbf{(3) Recursive $n$-layer meta (\metan{}).} \metan{} dissolves the dilemma rather than trading along it. The improver is frozen \emph{by design}, a single fixed LLM-prompted operation $\Omega$, and the recursion is applied to its \emph{input} instead. Because $\Omega$ consumes its own accumulated products (the traces of the stack below \emph{and} the code stack that produced them), each application operates on inputs of strictly higher order than the last. The operation itself never mutates, so it cannot destabilize the system. We do not fix depth in advance, but let it grow until $\Omega$ stops finding improvements (§\ref{sec:method:recursion}).

%------------------------------------------------------------------------------
\subsection{The Meta-Layer at Depth \texorpdfstring{$d$}{d}: \texorpdfstring{$\Omega$}{Omega} and \texorpdfstring{$M_d$}{M\_d}}
\label{sec:metalayer}
%------------------------------------------------------------------------------

\begin{figure}[t]
    \centering
    \begin{subfigure}[b]{0.60\linewidth}
        \centering
        \includegraphics[width=\linewidth]{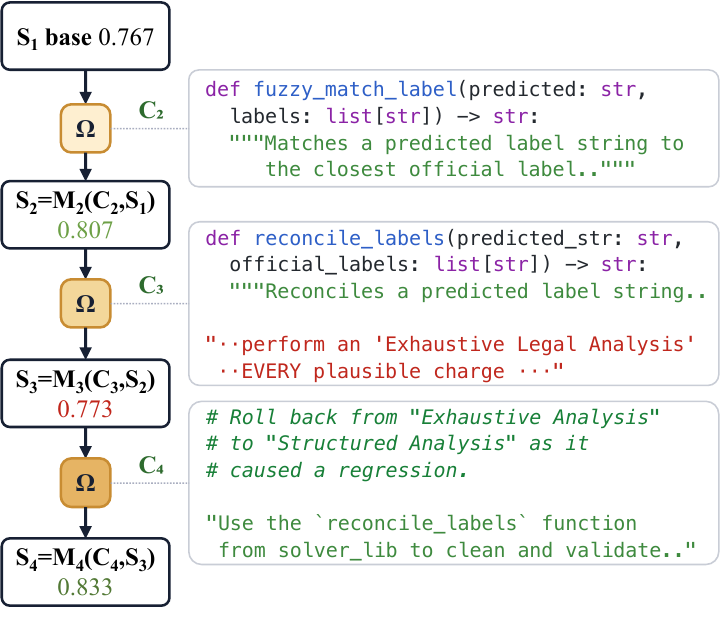}
        \caption{Build-step: $\Omega$ writes layers 2 to 4.}
        \label{fig:lawbench:build}
    \end{subfigure}
    \hfill
    \begin{subfigure}[b]{0.32\linewidth}
        \centering
        \includegraphics[width=\linewidth]{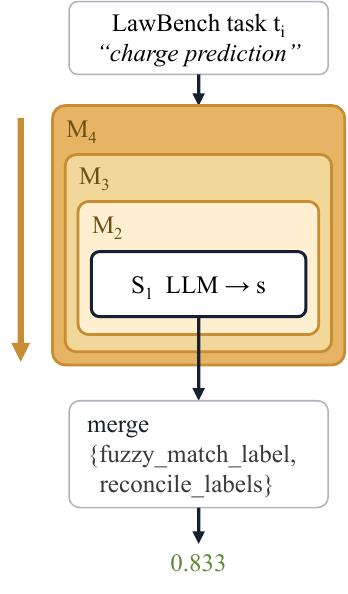}
        \caption{Run-step: $S_4$ executes the task.}
        \label{fig:lawbench:run}
    \end{subfigure}
    \caption{\textbf{The two phases of a meta-layer}, traced through a LawBench \emph{charge prediction} run (task score $\in [0,1]$). \textbf{(a)}~Build-step: each $\Omega$ call reads the traces of the stack below, and from depth 3 its code as well, then writes the next layer's code $C_d$. Depth 2 emits the helper \texttt{fuzzy\_match\_label}; depth 3 adds \texttt{reconcile\_labels} but couples it to an over-prescriptive \emph{``Exhaustive Legal Analysis''} directive that regresses the score; depth 4 rolls that directive back while keeping the helper. \textbf{(b)}~Run-step: the finished $S_4 = M_4 \!\circ\! M_3 \!\circ\! M_2 \!\circ\! S_1$ threads the task inward to the base solver and merges the accumulated helpers into the final script.}
    \label{fig:lawbench}
    \vspace{-5mm}
\end{figure}

A single meta-layer at depth $d$ unfolds in two phases, a \emph{build-step} that \emph{writes} the layer (executed once, offline) and a \emph{run-step} that \emph{executes} the layer (per task, online). Figure~\ref{fig:lawbench} traces both phases through a LawBench \emph{charge prediction} run (map a criminal-case description to the official charge labels), our running illustration for the rest of the section. Figure~\ref{fig:lawbench:build} shows $\Omega$ writing layers 2 to 4; Figure~\ref{fig:lawbench:run} shows the finished depth-4 stack executing a task. The build-step calls $\Omega$ to inspect the previous depth's traces and emit new code $C_d$; the run-step uses a wrapper $M_d$ to slot $C_d$ around the existing solver and run a task through the resulting stack.

\vspace{-3mm}
\paragraph{Build-step (writing the layer): the driver \texorpdfstring{$\Omega$}{Omega}.}
$\Omega$ is a fixed LLM-prompted procedure, i.e., one prompt template that does not change between depths or benchmarks. Given the previous depth's traces $\{\tau_i^{(d-1)}\}_{i=1}^{N}$, the stack of code earlier rounds emitted $[C_2, \dots, C_{d-1}]$, the task descriptions $\mathcal{T}$, and the current depth $d$, $\Omega$ writes a new piece of code:

\vspace{-5mm}

\begin{equation}
    \Omega : \big(\{\tau_i^{(d-1)}\}_{i=1}^{N},\; [C_2, \dots, C_{d-1}],\; \mathcal{T},\; d\big) \;\longmapsto\; C_d.
    \label{eq:omega}
\end{equation}
The output $C_d = (f_{\text{pre}}^{(d)}, \mathcal{L}^{(d)})$ is a pair. The \emph{pre-process} $f_{\text{pre}}^{(d)}$ is a small Python function that runs before each solver call and injects strategic context. In Figure~\ref{fig:lawbench:build}, the depth-3 $f_{\text{pre}}^{(3)}$ demands an \emph{``Exhaustive Legal Analysis''} of every plausible charge, while the depth-4 version instead tells the solver to clean its labels with \texttt{reconcile\_labels}. The \emph{code library} $\mathcal{L}^{(d)}$ holds reusable Python functions the solver may call, such as \texttt{fuzzy\_match\_label()} emitted at depth 2 and \texttt{reconcile\_labels()} added at depth 3. Concretely, $\Omega$'s response contains a \texttt{rationale} block, a \texttt{pre\_process} block, and zero or more \texttt{solver\_lib:<name>} blocks, one per library function (full depth-2 example in Appendix~\ref{app:depth2_code}).

Holding $\Omega$ fixed makes the design principle of §\ref{sec:method:paradigms} operational, since any gain at depth $d{>}2$ must then come from richer \emph{input} rather than a changed improver. The only part that adapts with depth is the trace summary $\Omega$ ingests. Its formatting depends on depth, and from depth 3 upward it shifts attention from individual errors to patterns across the stack (details in Appendix~\ref{app:trace_payload}). This formatting partly scaffolds the tactical-versus-strategic \emph{division of attention} analyzed in §\ref{sec:experiments}, but it only shapes what each layer \emph{sees}; no prompt prescribes the code a layer writes.

\vspace{-3mm}
\paragraph{Run-step (executing the layer): the wrapper \texorpdfstring{$M_d$}{M\_d}.}
When task $t_i$ comes in at depth $d$, the wrapper $M_d$ slots $C_d$ around the previous solver $S_{d-1}$ in five stages, sketched in Figure~\ref{fig:lawbench:run}. On the way in, (1)~the outermost $f_{\text{pre}}^{(d)}$ runs first and produces a strategic-context string $\text{ctx}_d$ for this task, and (2)~$\text{ctx}_d$ threads inward through the pre-processes of the layers below, each refining it. On the way out, (3)~the base solver sees the merged contexts and returns a script $s$, (4)~the union code library $\mathcal{L}^{(2)} \!\cup\! \cdots \!\cup\! \mathcal{L}^{(d)}$ is prepended to $s$, with deeper layers overriding by name on collisions, and (5)~the result is sandbox-executed to produce a depth-$d$ trace $\tau_i^{(d)}$. Because $M_d$ does not touch the inner solver's internals, wrappers compose cleanly, so the depth-$n$ solver is just nested calls bottoming out at the base solver $S_1$:
\begin{equation}
    S_d \;=\; M_d(C_d, S_{d-1}) \;=\; M_d \circ M_{d-1} \circ \cdots \circ M_2 \circ S_1.
    \label{eq:wrapper-compose}
\end{equation}
$M_d$ never mutates the inner solver, the task object, or earlier layers' libraries, and $M_d$ statically validates and smoke-tests injected code in an isolated namespace before prepending it (rationale and safeguards in Appendix~\ref{app:design_rationale}).

%------------------------------------------------------------------------------
\subsection{Compositional Recursion: Strategies, Tactics, and What \texorpdfstring{$\Omega$}{Omega} Sees}
\label{sec:method:recursion}
%------------------------------------------------------------------------------

Recursion buys two things over a flat improver, \emph{multiplicative coverage} of the strategy/tactic space and \emph{strictly richer input} to the improver itself. The first is about the behaviors a $d$-layer stack can express when it runs; the second is about what $\Omega$ has to look at when it writes the next layer.

\vspace{-3mm}
\paragraph{Strategies condition tactics.}
During the run-step, each pre-process receives the context string of the layer above and emits its own for the layer below, $f_{\text{pre}}^{(d-1)}(t_i,\, \text{ctx}_d) \rightarrow \text{ctx}_{d-1}$. Layers therefore condition one another; the context a layer emits is shaped by the context threaded down from above, so a deeper layer sets a frame that shallower layers fill in. With $k_d$ distinct behaviors at depth $d$, conditioning admits up to $\prod_{d=2}^{n} k_d$ joint configurations, against the $\sum_d k_d$ available to an unconditioned flat architecture with the same behaviors (at $n{=}4$ and $k_d{=}3$, 27 vs.\ 9). This is an \emph{upper bound on expressible configurations}, not a measured count. Its measured signature appears in §\ref{sec:experiments} as the gap between archive-best and best-single-chain scores, together with an ablation attributing most of the recursion gain to this conditioning channel. Figure~\ref{fig:lawbench:run} shows the conditioning at work. The outermost layer runs first and sets the strategic frame, and each inner layer refines the guidance within that frame. The base solver then receives the merged contexts of all layers together with the union library.

Conditioning also means layers can \emph{interfere}. A deeper layer's prescription can override useful shallower guidance, and we observe such per-task regressions in practice (quantified in §\ref{sec:experiments}). The running example is one such cycle. In Figure~\ref{fig:lawbench:build}, the depth-3 directive regresses the score from $0.807$ to $0.773$ even though the helper shipped alongside it is sound. Because the depth-4 $\Omega$ reads the code stack $[C_2, C_3]$ as well as the traces (Eq.~\ref{eq:omega}), it can attribute the regression to the directive rather than the helper. Its rationale states the rollback explicitly (\emph{``roll back from Exhaustive Analysis to Structured Analysis as it caused a regression''}), and its $C_4$ restores the earlier strategy while keeping \texttt{reconcile\_labels}, recovering to $0.833$. Interference is thus a cost of expressiveness rather than a bug in it, and it is repaired at two levels, within a chain by $\Omega$ itself, as here, and across chains by the archive and consolidation mechanisms of §\ref{sec:method:orchestration}.

\vspace{-3mm}
\paragraph{What $\Omega$ sees vs.\ what flat self-refinement sees.}
A flat self-refinement loop at iteration $j$ observes only the traces of its first $j{-}1$ iterations, an accumulating log of \emph{what happened} that never includes the code \emph{that made it happen}. By construction, $\Omega$ at depth $d \geq 3$ sees both the previous depth's traces \emph{and} the code stack $[C_2, \dots, C_{d-1}]$ that produced them. This enables higher-order reasoning that a flat loop cannot perform however many iterations it runs, such as the rollback in Figure~\ref{fig:lawbench:build} (\emph{``$C_3$'s directive over-constrained the analysis; roll it back at depth 4''}). Whenever $C_{d-1}$ produces observable changes, $\Omega$'s information at depth $d$ is a strict superset of its information at depth $d{-}1$. The fixed-driver systems of §\ref{sec:method:paradigms} sit between the two. Their improver sees only what their editable layers produced, a strict subset of what \metan{}'s $\Omega$ sees. \metan{} thus reframes the design problem. \textbf{The gain comes from giving $\Omega$ more to read, not from rewriting $\Omega$ itself.} §\ref{sec:experiments} isolates this claim empirically; removing recursion while leaving the orchestration unchanged lowers scores on every backbone and benchmark we ablate, and the drop is largest where the base solver is weakest.

%------------------------------------------------------------------------------
\subsection{Orchestration: Linear and Evolutionary Modes}
\label{sec:method:orchestration}
%------------------------------------------------------------------------------

Two orchestrators grow the stack on top of $\Omega$ and $M_d$, a simple linear deepener and an evolutionary archive search.

\vspace{-3mm}
\paragraph{Linear meta-recursion.}
The simplest orchestrator greedily deepens the stack one layer at a time. Starting from the base solver $S_1$, at each step $\Omega$ observes the current traces and produces a new $C_d$, which the wrapper $M_d$ composes onto $S_{d-1}$ to form $S_d$. The loop stops when (a)~$\Omega$ returns empty code, (b)~$P$ consecutive layers fail to improve the mean score by more than a scale-aware tolerance $\epsilon \cdot R$ ($R$ is the score range frozen at run start), or (c)~a max depth $D$ is reached. Depth is therefore set by convergence rather than prescribed in advance; in practice no reported run reaches the cap.

\vspace{-3mm}
\paragraph{Evolutionary archive.}
Linear recursion is fragile to one unlucky $\Omega$ call; a single poor injection ends the chain even if a different injection at the same depth would have unlocked further progress. The evolutionary variant (Algorithm~\ref{alg:evolutionary}, Appendix~\ref{app:orchestrator}) hedges by maintaining a \emph{monotonically growing} archive $\mathcal{A}$ of candidate chains and searching over it instead of greedily extending one. Each iteration samples $B$ parents with weight $w(c) \propto \bar S(c) + \alpha / (1 + \text{children}(c))$, so high scores are preferred and the $\alpha$ term acts as an exploration bonus favoring chains extended less often. Each parent yields $K$ children, with $\Omega$'s sampling temperature cycled for diversity; on any task where the parent underperforms the archive, \emph{cross-candidate inspiration} appends the best rival trace to $\Omega$'s prompt. The archive also tracks the best score any chain achieves on each task; the mean of these per-task bests is the \emph{archive-best} $\bar S^\star$, reported alongside the best-single-chain score in every results table. The distance between them is what the archive gains by letting different tasks be won by different chains. The loop terminates after $P$ consecutive non-improving iterations.

\vspace{-3mm}
\paragraph{Consolidation guard.}
Because deeper layers can regress individual tasks (§\ref{sec:method:recursion}), the orchestrator also supports a \emph{consolidation} mode in which each candidate targets a single focus task while inheriting the archive's frozen best traces for all others, making the per-task-best trajectory monotone \emph{by construction}. We claim this mode for its zero-regression guarantee rather than for mean lift (study and compute-matched control in Appendix~\ref{app:consolidation}).

\vspace{-3mm}
\paragraph{Agentic solver.}
Both orchestrators can wrap a single-shot Layer-1 solver or an \emph{agentic} solver that runs a generate, execute, observe, refine loop of up to $T$ turns ($T\!=\!8$); the agentic solver receives the same pre-process and code library and returns the best-scoring script. Comparing the two under identical evolutionary hyperparameters lets us separate the contribution of within-task iteration from that of $\Omega$'s cross-task signal.

\noindent\textbf{Design decisions.} We document the remaining choices (same model at all layers; wrapper-not-patching; 3:1 failure-biased trace ratio; per-benchmark $\epsilon$, patience, and beam settings) in Appendix~\ref{app:design_rationale}.

%% file: sections/3_experiments.tex
%==============================================================================
\section{Experiments}
\label{sec:experiments}
%==============================================================================
\label{sec:main_results}

\paragraph{Setup.}
\label{sec:exp_setup}
We report results on two backbones, Gemma~4 31B-IT and GPT-5.2; unless noted otherwise, every number is a mean $\pm$ stdev over three seeds (42, 43, 44). Eight benchmark families span three solver substrates: \emph{Python source} (CO-Bench~\citep{cobench2026}, 36 NP-hard problems; AlphaEvolve Math~\citep{alphaevolve2025}; Symbolic Regression (SR), 4 domains; AlgoTune, 8 tasks; ARC-AGI-2~\citep{chollet2026arcagi2newchallengefrontier}, 120 tasks), \emph{bash inside a Docker sandbox} (TerminalBench~2.0 (TB2), 89 hard tasks across 13 categories), and \emph{prompt rewrite for a fixed downstream model} (Symptom2Disease (S2D) and LawBench charge prediction). Two \metan{} variants run under identical evolutionary orchestration with the single $\Omega$ template of \S\ref{sec:metalayer}, \emph{single-shot} (one LLM call per task; isolates $\Omega$'s contribution) and \emph{agentic} (observe-act loop, max 8 turns per task); we select $B$, $K$, patience, and max-iter per benchmark (Appendix~\ref{app:setup_details}, Table~\ref{tab:hparams}). We deviate from the seed policy in two places. TB2 stdevs are computed across the 13 task categories rather than across seeds (TB2 under GPT-5.2 is single-seed, s42), and benchmarks without a held-out split (AlphaEvolve Math, AlgoTune, SR) report the benchmark score directly. Full configurations and costs are in Appendix~\ref{app:setup_details}.

\paragraph{Baselines and estimators.}
G\"odel Agent (GA)~\citep{godelAgent2024} and OpenEvolve (OE)~\citep{alphaevolve2025, openevolve} are the prior self-improving comparators, run with the same base model on the benchmarks marked in Tables~\ref{tab:vs_baselines} and~\ref{tab:vs_baselines_gpt52}. We report GA in a corrected \emph{per-task} configuration, since its published single-solver interface collapses structurally on CO-Bench; Appendix~\ref{app:godel_details} documents both configurations. OE is a per-artifact evolutionary comparator (island/MAP-Elites) run at near-identical token budget on CO-Bench; \S\ref{sec:parity} equalizes compute explicitly where the default budgets differ. For \metan{} we report both estimators everywhere, \emph{archive-best} ($\bar S^\star$ of \S\ref{sec:method:orchestration}) and \emph{best single chain}. Their difference is what per-task selection buys over committing to a single chain (\S\ref{sec:method:recursion}).

% \vspace{-2mm}
\subsection{Main Results}
\vspace{-2mm}
\begin{table}[ht]
\caption{\textbf{\metan{} agentic vs.\ prior self-improving agents (Gemma~4 31B-IT, mean $\pm$ stdev over seeds 42/43/44).} Held-out test for CO-Bench, S2D, and LawBench; benchmark score for AlphaEvolve Math, AlgoTune (speedup $\times$), and SR (4-domain mean). ARC-AGI-2 and the SR baselines were run under GPT-5.2 only (Table~\ref{tab:vs_baselines_gpt52}); n/r = not run.}
\label{tab:vs_baselines}
\centering
\small
\setlength{\tabcolsep}{3pt}
\resizebox{0.98\columnwidth}{!}{%
\begin{tabular}{@{}lcccccc@{}}
\toprule
\textbf{Method} & \textbf{CO-Bench} & \textbf{S2D} & \textbf{LawBench} & \textbf{AE Math} & \textbf{AlgoTune} & \textbf{SR} \\
\midrule
\textbf{\metan{}} archive-best & \textbf{0.851 $\pm$ 0.014} & 0.733 $\pm$ 0.015 & \textbf{0.815 $\pm$ 0.013} & \textbf{0.869 $\pm$ 0.045} & \textbf{$\times$15.10 $\pm$ 2.4} & \textbf{5.22 $\pm$ 0.38} \\
\textbf{\metan{}} best chain   & 0.782 $\pm$ 0.016 & \textbf{0.743 $\pm$ 0.034} & 0.796 $\pm$ 0.046 & n/r & n/r & 4.68 $\pm$ 0.45 \\
OpenEvolve~\citep{openevolve}  & 0.814 $\pm$ 0.022 & 0.718 $\pm$ 0.022 & 0.745 $\pm$ 0.034 & 0.802 $\pm$ 0.052 & $\times$10.45 $\pm$ 1.8 & n/r \\
G\"odel Agent~\citep{godelAgent2024} & 0.451 $\pm$ 0.023 & 0.710 $\pm$ 0.034 & 0.775 $\pm$ 0.023 & 0.581 $\pm$ 0.061 & $\times$13.22 $\pm$ 2.7 & n/r \\
\bottomrule
\end{tabular}%
}
\vspace{-2mm}
\end{table}
\begin{table}[ht]
\caption{\textbf{Cross-model condition: GPT-5.2 backbone, mean $\pm$ stdev over seeds 42/43/44.} Held-out test for CO-Bench and S2D; benchmark score for AlphaEvolve Math, SR (4-domain mean), and ARC-AGI-2 (dev score; the held-out test result is reported in the text). LawBench and AlgoTune were not run under GPT-5.2 for budget reasons; their Gemma columns appear in Table~\ref{tab:vs_baselines}. n/r = estimator not run for this cell.}
\label{tab:vs_baselines_gpt52}
\centering
\small
\setlength{\tabcolsep}{3pt}
\resizebox{0.9\columnwidth}{!}{%
\begin{tabular}{@{}lccccc@{}}
\toprule
\textbf{Method} & \textbf{CO-Bench} & \textbf{S2D} & \textbf{AE Math} & \textbf{SR} & \textbf{ARC-AGI-2} \\
\midrule
\textbf{\metan{}} archive-best & \textbf{0.870 $\pm$ 0.011} & 0.725 $\pm$ 0.014 & \textbf{0.917 $\pm$ 0.016} & \textbf{5.03 $\pm$ 0.20} & \textbf{0.331 $\pm$ 0.010} \\
\textbf{\metan{}} best chain   & 0.806 $\pm$ 0.017 & \textbf{0.734 $\pm$ 0.020} & 0.820 $\pm$ 0.024 & n/r & 0.123 (1 seed) \\
OpenEvolve~\citep{openevolve}  & 0.702 $\pm$ 0.025 & 0.721 $\pm$ 0.007 & 0.726 $\pm$ 0.046 & 3.70 $\pm$ 0.13 & 0.003 $\pm$ 0.001 \\
G\"odel Agent~\citep{godelAgent2024} & 0.527 $\pm$ 0.033 & 0.708 $\pm$ 0.037 & 0.674 $\pm$ 0.068 & 2.45 $\pm$ 1.63 & 0.054 $\pm$ 0.006 \\
\bottomrule
\end{tabular}%
}
\vspace{-4mm}
\end{table}

\paragraph{Headlines.}
\metan{} agentic leads every benchmark family in Tables~\ref{tab:vs_baselines} and~\ref{tab:vs_baselines_gpt52} on at least one estimator, under both backbones, though the size of the margin varies by benchmark. On the code substrates the wins are large and stable. The CO-Bench margin over OpenEvolve reaches $+0.168$ on GPT-5.2 with per-seed ranges that do not overlap, AlphaEvolve Math is ahead by a similar margin, and SR leads on all four domains. On ARC-AGI-2 the gap is categorical rather than quantitative, with both baselines near the floor. On the prompt-rewrite benchmarks the margins shrink to a few points. \metan{} still leads S2D on the mean under both backbones, though the per-seed ranges overlap, and LawBench separates by $+0.040$ over G\"odel Agent. Two architectural levers explain these wins, recursion across depths and cross-task transfer through code-library injection.

\vspace{-3mm}
\paragraph{ARC-AGI-2: the categorical case.}
ARC-AGI-2 was designed to probe fluid intelligence; its tasks cannot be solved by refining a known skill, only by abstracting new ones, so it separates the two levels of operation. Object-level iteration stays near the floor, and even \metan{}'s best single chain reaches only $0.123$, but the full meta-level stack, in which $\Omega$ abstracts transform primitives from traces and the archive composes them across tasks, reaches $0.331$. On a benchmark designed to resist skill memorization, the gains come from the meta level rather than from better object-level search.

\vspace{-3mm}
\paragraph{CO-Bench.}
The archive recovers tasks that a single LLM call cannot solve at all. On a representative run (Gemma, s42), constrained guillotine cutting goes $0.000\!\to\!0.996$ and maximal independent set $0.000\!\to\!0.908$. In aggregate (Table~\ref{tab:vs_baselines}), archive-best leads OpenEvolve on Gemma and widens the margin to $+0.168$ on GPT-5.2 with disjoint per-seed ranges; the ordering of the three systems is the same on both backbones, so the gap is not a backbone artefact. The archive also clears its own best single chain by $0.06$ to $0.07$ on both backbones, the empirical signature of the multiplicative-coverage bound of \S\ref{sec:method:recursion}.

\vspace{-3mm}
\paragraph{Mathematical and algorithmic discovery.}
On AlphaEvolve Math, $\Omega$ injects two callable primitives, \texttt{simulated\_annealing()} and \texttt{basin\_hopping()}, that turn an unsolvable seed into a strong solver; on the Gemma s42 run the seed climbs from $0.222$ to $0.709$ single-shot and from $0.435$ to $0.883$ agentic. Across seeds \metan{} leads both baselines under either backbone, and SR shows the same pattern on all four domains (per-domain breakdown in Appendix Table~\ref{tab:sr_per_domain}). AlgoTune is the one benchmark where the agentic variant underperforms its single-shot sibling ($\times14.11$ vs.\ $\times18.47$ on s42), or $\times15.96$ vs.\ $\times18.47$ on the seven tasks both variants logged (Appendix Table~\ref{tab:algotune_per_task}), though the three-seed agentic mean ($\times15.10$) still beats every baseline. The benchmark's pre-optimized kernel contract is already extracted by the seed, $\Omega$'s extra context over-constrains, and the per-task diagnostic in Appendix~\ref{app:algotune} makes this mechanism testable.

\vspace{-3mm}
\paragraph{TerminalBench 2.0.}
Running either baseline on the Docker substrate would require substantial changes to its code, so TB2 measures \metan{} against its own seeds instead. The two variants converge to similar absolute gains from very different starting points, so $\Omega$'s contribution is roughly orthogonal to seed strength. On Gemma, single-shot lifts the 13-category mean by $+0.21$ from a weak $0.067$ seed and agentic by $+0.23$ from a $0.258$ seed; on GPT-5.2 agentic reaches $0.634$. The archive-over-chain margin holds here too, and the benchmark exercises language-adaptive injection, with $\Omega$ emitting both bash functions and Python helper scripts across the 13 sandbox categories.

\vspace{-3mm}
\paragraph{Text classification.}
The prompt-rewrite benchmarks bound what any meta-layer can add, since the action surface is a single prompt over a fixed label set. On S2D the best chain leads both baselines under either backbone, by $+0.025$ (Gemma) and $+0.013$ (GPT-5.2), though the per-seed ranges overlap and we do not claim significance. LawBench separates further, with archive-best ahead of G\"odel Agent by $+0.040$ and OpenEvolve by $+0.070$. The depth progression stays legible even here. On a representative S2D chain, $\Omega$ moves from a CoT pipeline (d2) to a \texttt{robust\_label\_match} helper (d3) to a \texttt{classification\_prompt\_builder} (d4); on LawBench, d2 switches to Chinese prompts, d3 over-constrains and regresses, and d4 rolls the prompt intensity back, the same regression-then-rollback cycle analyzed in \S\ref{sec:method:recursion}.

% Panels generated by figures/make_figure4.py (data sources documented there).
\begin{figure}[t]
  \centering
  \begin{subfigure}[b]{0.245\linewidth}
    \centering
    \includegraphics[width=\linewidth]{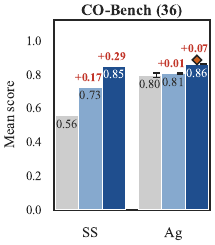}
  \end{subfigure}\hfill
  \begin{subfigure}[b]{0.245\linewidth}
    \centering
    \includegraphics[width=\linewidth]{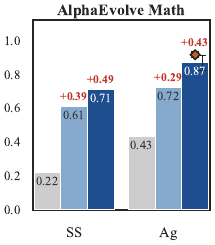}
  \end{subfigure}\hfill
  \begin{subfigure}[b]{0.245\linewidth}
    \centering
    \includegraphics[width=\linewidth]{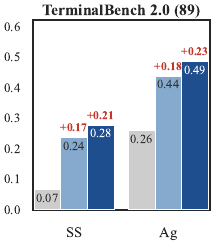}
  \end{subfigure}\hfill
  \begin{subfigure}[b]{0.245\linewidth}
    \centering
    \includegraphics[width=\linewidth]{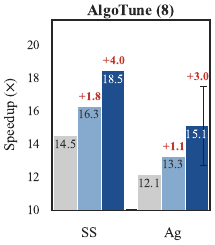}
  \end{subfigure}
  \vspace{0.3em}
  \includegraphics[width=0.97\linewidth]{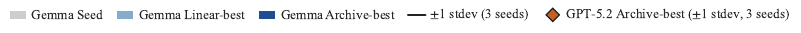}
  \vspace{-2mm}
  \caption{\textbf{\metan{} variants, single-shot vs.\ agentic,} on four representative benchmarks (Gemma~4 31B-IT; hyperparameters in Table~\ref{tab:hparams}). SS = single-shot, Ag = agentic. Each group shows the seed (the Layer-1 solver, before any meta-layer), linear-best (best single chain), and archive-best (per-task best across the archive), annotated with the raw score and the gain over that group's seed. Panels plot validation or benchmark scores, with held-out test results in Table~\ref{tab:vs_baselines}; AlgoTune is speedup ($\times$). Black whiskers mark three-seed means $\pm 1$ stdev (seeds 42/43/44), and orange diamonds the three-seed GPT-5.2 archive-best where that backbone was run.}
  \label{fig:variants}
  \vspace{-5mm}
\end{figure}

\vspace{-3mm}
\paragraph{Single-shot vs.\ agentic.}
In Figure~\ref{fig:variants}, the linear-best bar isolates the recursion contribution and the residual to archive-best isolates what the archive itself adds. Agentic improves archive-best on seven of the eight benchmark families, with the largest gains on TB2, AlphaEvolve Math, and SR. The agentic seed itself rises substantially (TB2 $0.067\!\to\!0.258$, AlphaEvolve $0.222\!\to\!0.435$), so the inner observe-act loop extracts a stronger Layer-1 baseline on top of which $\Omega$ adds further gains. Token cost is roughly 4 to $10\times$ higher in agentic mode (8 calls per task), trading compute for richer trace signal; AlgoTune is the exception where the richer signal over-constrains (\S\ref{sec:case_study}).

In single-seed Gemma runs, agentic \metan{} reaches higher scores in fewer evolution iterations than both baselines on CO-Bench, LawBench, and S2D; single-shot trails both on S2D (Appendix~\ref{app:convergence}).

\vspace{-2mm}
\subsection{Baseline Parity and Diagnostics}
\label{sec:parity}
\vspace{-1mm}

All comparisons use the same backbone within each condition, and the CO-Bench tables report GA's corrected per-task configuration. \metan{} and OE use near-identical token budgets on CO-Bench. On Gemma prompt benchmarks, capping \metan{} at OE's token budget yields S2D $0.732 \pm 0.023$ vs.\ $0.718 \pm 0.022$ and LawBench $0.784 \pm 0.013$ vs.\ $0.745 \pm 0.034$; the S2D mean $\pm$ standard-deviation intervals overlap. On GPT-5.2 CO-Bench, increasing GA's evolution budget $10\times$ (seed 42) reaches $0.628$. Even corrected, the residual gap to \metan{} is architectural rather than budgetary. Appendix~\ref{app:oe_budget_controls} gives the OE budget controls and evaluation counts; Appendix~\ref{app:godel_details} documents GA's configuration and full budget sweep.

\vspace{-2mm}
\subsection{What Each Depth Learns}
\label{sec:case_study}
\vspace{-1mm}

\begin{wrapfigure}[24]{r}{0.48\textwidth}
\centering
\includegraphics[width=\linewidth]{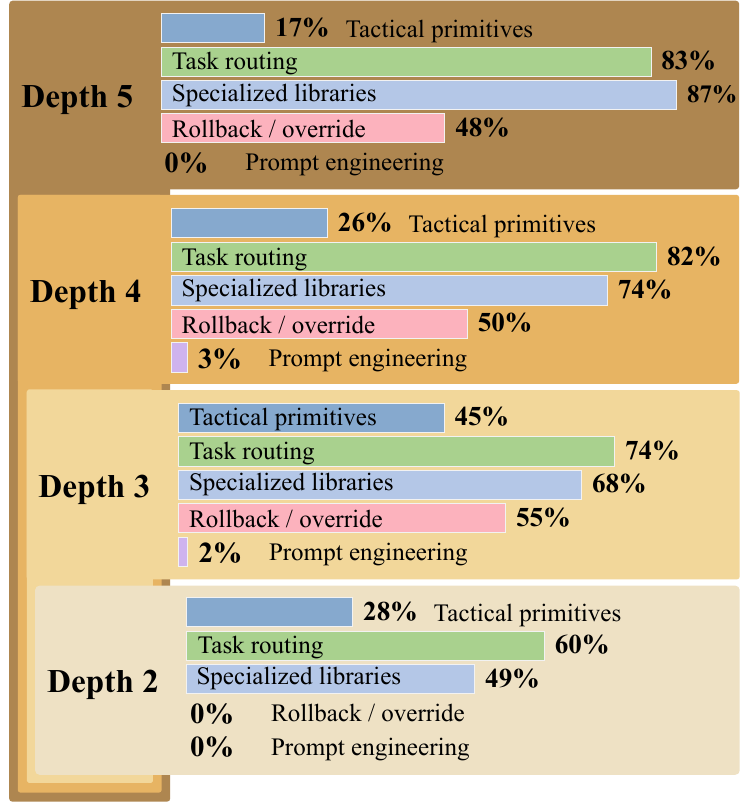}
\caption{\textbf{Emergent roles on code benchmarks.} Fractions require both raters' agreement; categories are non-exclusive.}
\label{fig:emergent_roles}
\end{wrapfigure}

Figure~\ref{fig:emergent_roles} shows distinct roles although $\Omega$'s prompt prescribes none, across 288 code emissions from CO-Bench, AlphaEvolve Math, and SR. Two independently prompted LLM raters (GPT-5.2 and Kimi-K2.6) annotate 596 emissions over seven categories across both substrates; we count only labels on which both agree. On code benchmarks, rollback is absent at depth 2 and appears at depth 3 (55\%); tactical primitives peak at depth 3 (45\%) and fall to 17\% by depth 5 as specialized libraries take over. Prompt benchmarks remain dominated by prompt engineering, with tactical primitives never exceeding 9\% (full comparison in Appendix~\ref{app:patterns_rubric}, Figure~\ref{fig:emergent_roles:full}). Agreement is uneven: mean Cohen's $\kappa=0.59$, with weaker agreement on tactical primitives ($0.37$) and compositional architectures ($0.30$), warranting caution about abstract role labels.

Depth also introduces interference: 41\% of (chain, task) pairs regress at depth 3. Nevertheless, depth-${\geq}4$ candidates win 31\% of CO-Bench tasks, 80\% of SR-matsci, and 70\% of SR-phys-osc, so the archive preserves useful shallow fallbacks alongside deep specialists. Gains depend on failure-mode diversity and headroom above the seed; prompt-only tasks offer a narrower action surface, and AlgoTune illustrates how extra context can over-constrain a strong seed. Appendix~\ref{app:depth_analysis} gives the task-level trajectories, substrate comparisons, and boundary cases. The consolidation guard freezes non-focus tasks to prevent regressions; its study and compute-matched control are reported in Appendix~\ref{app:consolidation}.

\vspace{-2mm}
\Needspace{6\baselineskip}
\subsection{Ablations}
\label{sec:ablations}
\vspace{-1mm}

\begin{wraptable}[13]{R}{0.52\textwidth}
\vspace{-1mm}
\centering
\setlength{\tabcolsep}{3pt}
\caption{\textbf{Component ablations.} Each row removes one mechanism and keeps the rest, and comparisons are like-for-like within a block.}
\label{tab:ablations}
\small
\resizebox{\linewidth}{!}{%
\begin{tabular}{@{}llcc@{}}
\toprule
\textbf{Condition} & \textbf{Removal} & \textbf{Score} & \textbf{$\Delta$} \\
\midrule
\metan{} (full; Gemma, CO) &                            & 0.845 &  \\
$-$code library        & $\mathcal{L}^{(d)}$ injection  & 0.825 & $-0.020$ \\
$-$outer-context       & inter-layer ctx$_d$            & 0.751 & $-0.094$ \\
$-$recursion (depth-1) & meta-layer stacking            & 0.714 & $-0.131$ \\
\midrule
\metan{} (full; GPT-5.2, CO) &                          & 0.886 &  \\
$-$recursion (depth-1) & meta-layer stacking            & 0.806 & $-0.080$ \\
\midrule
\metan{} (full; GPT-5.2, AE) &                          & 0.917 &  \\
$-$recursion (depth-1) & meta-layer stacking            & 0.759 & $-0.158$ \\
\bottomrule
\end{tabular}%
}
\end{wraptable}

\textbf{Conditioning is the dominant channel.} Removing the inter-layer context, a single string passed between layers, explains ${\sim}72\%$ of recursion's gain over a depth-1 baseline. The richer code-library channel (callable Python functions) explains ${\sim}15\%$; the remaining ${\sim}13\%$ comes from the recursion machinery itself (beam, retry, inspiration) (Table~\ref{tab:ablations}).

\paragraph{Recursion vs.\ flat alternatives.}
The baselines of \S\ref{sec:parity} double as architecture-level ablations, each removing a different piece of the architecture. GA removes recursion in favour of flat self-modification. Even in its corrected configuration and at $10\times$ budget it plateaus at $0.628$ on GPT-5.2 CO-Bench, well short of \metan{}'s $0.870$, which makes the margin a limit of GA's architecture rather than of its compute. OE keeps a search loop but runs it per task, so its CO-Bench test gap on both backbones (Tables~\ref{tab:vs_baselines} and~\ref{tab:vs_baselines_gpt52}) is what cross-task transfer is worth, measured at matched compute on CO-Bench and at capped compute on the prompt benchmarks (\S\ref{sec:parity}). Inside \metan{}, the depth-1 ablation removes only the recursion. Its lift reproduces across backbones and benchmarks ($+0.131$ Gemma CO-Bench, $+0.080$ GPT-5.2 CO-Bench, $+0.158$ GPT-5.2 AE Math) and grows as the inner-loop ceiling drops, which is what \S\ref{sec:method:recursion} predicts.

\vspace{-3mm}
\paragraph{Where the depth gain lives.}
The ablation decomposition says the upper layer's strategy string reshapes the lower layer's behavior at every step, and that this conditioning carries more of the gain than the callable code the layer also hands down. This could be seen as the empirical form of the $\prod_d k_d$ argument of \S\ref{sec:method:recursion}. Depth multiplies coverage by conditioning, the archive turns that coverage into per-task wins (the archive-over-chain margins of Tables~\ref{tab:vs_baselines} and~\ref{tab:vs_baselines_gpt52}), and the consolidation study (Appendix~\ref{app:consolidation}) shows those wins can be collected monotonically.

%% file: sections/4_conclusion.tex
%==============================================================================
\vspace{-3mm}
\section{Conclusion}
\label{sec:conclusion}
%==============================================================================

We presented \metan{}, a self-improving agent built by holding one meta-operation $\Omega$ fixed and recursing on its \emph{input} rather than on the operation itself. Each application of $\Omega$ reads the traces of the stack below together with the code that produced them, so it starts from a strictly larger information set than the application before it, and the stack deepens until improvements stop. This dissolves the dilemma that shapes prior work, where recursing the improver buys depth only by putting stability at risk, and it does so without ever editing the improver.

The evidence supports the architecture on both axes it was tested against. Across eight benchmark families and two backbones, \metan{} leads every benchmark family on at least one estimator, with the widest margins where a benchmark fails in many distinct ways. On ARC-AGI-2, built to resist skill memorization, object-level iteration stays near the floor while the meta-level stack reaches $0.331$. The component ablation locates the gain more precisely than expected. Most of what recursion buys comes from the plainest channel between layers, a passed string of context, which accounts for roughly $72\%$ of the lift, while callable code transfer accounts for about $15\%$. Roles then emerge across depth without any prompt prescribing them explicitly.

\vspace{-3mm}
\paragraph{Limitations and Future Work.}
We run the same model at the base solver and at every $\Omega$ invocation. That is the right control for separating depth gains from capability gains, but it leaves the practical deployment case untested, in which a stronger model sits at $\Omega$ over a weaker base; whether depth keeps paying when only the improver is upgraded is the natural next experiment. Also, since conditioning rather than code transfer carries most of the lift, we could further enrich the context passed between layers, for instance, by replacing its free-form string with a structured or typed representation. Another direction would be investigating how deep the stack could usefully go. We observe that runs stop between depth 3 and depth 6 because $\Omega$ stops finding improvements, not because the base model runs out of reasoning capacity at higher meta-levels or the context fills with accumulated layer code.

%% file: sections/9_appendix.tex
%==============================================================================
\section{Notation}
\label{app:notation}
%==============================================================================

Table~\ref{tab:glossary} collects the symbols used in \S\ref{sec:method}.

\begin{table}[h]
\centering
\caption{\textbf{Symbol glossary for §\ref{sec:method}.}}
\label{tab:glossary}
\begin{tabular}{@{}ll@{}}
\toprule
\textbf{Symbol} & \textbf{Plain English} \\
\midrule
$\Omega$ & fixed meta-operation, called at each build-step \\
$M_d$,\, $S_d$ & depth-$d$ wrapper and resulting solver \\
$C_d$ & $\Omega$'s output: pre-process $f_{\text{pre}}^{(d)}$ + library $\mathcal{L}^{(d)}$ \\
$\tau_i^{(d)}$ & execution trace at depth $d$ on task $i$ \\
$\text{ctx}_d$ & outer-context passed from layer $d$ to layer $d{-}1$ \\
\bottomrule
\end{tabular}
\end{table}

%==============================================================================
\section{Related Work}
\label{app:related_work_full}
%==============================================================================

\paragraph{Single-level self-improvement and optimization.}
Single-feedback-loop systems (Self-Refine~\citep{selfrefine2023}, Reflexion~\citep{reflexion2023}, Self-Debugging~\citep{selfdebug2024}, RISE~\citep{rise2024}) generate, evaluate, reflect, retry. Evolutionary approaches maintain populations. FunSearch~\citep{funsearch2024} pairs a frozen LLM with an evaluator in an island-based loop, AlphaEvolve~\citep{alphaevolve2025} and OpenEvolve~\citep{openevolve} generalize this to entire codebases, Evolution of Heuristics~\citep{eoh2024} co-evolves natural-language ``thoughts'' and code, and ReEvo~\citep{reevo2024} introduces ``verbal gradients.'' Architecture-search systems (ADAS~\citep{adas2025}, AgentSquare~\citep{agentsquare2025}, AFlow~\citep{aflow2025}) program new agents from an archive or run MCTS over code-workflows. All operate at a \emph{single level of abstraction}, in which the mechanism that generates, mutates, or searches is itself fixed; the critic is never critiqued, and the evolutionary loop never evolves its own selection rule. \metan{} lifts this restriction by stacking meta-layers in which each layer generates code that modifies the layer below and is itself evaluated by the layer above; stack depth is set by convergence.

\paragraph{Self-referential agents.}
Schmidhuber's G\"odel Machine~\citep{schmidhuber2003} provides the theoretical foundation for provably self-improving agents (formal proof requirements limit practicality). STOP~\citep{stop2024} demonstrates recursive scaffolding improvement but realizes a single meta-step. G\"odel Agent~\citep{godelAgent2024} rewrites its own source code; the Darwin G\"odel Machine~\citep{darwinGodel2025} combines evolutionary population maintenance with self-referential modification, reaching 50\% on SWE-bench; HyperAgents~\citep{hyperagents2026} merges task and meta agents into a single editable program. EvoX~\citep{liu2026evox} extends evolutionary program search to a meta-evolution regime in which the search operators themselves evolve. All empirically realize at most $\sim$2.5 meta-levels because at least one driver layer is held fixed for stability. G\"odel Agent's action API and goal prompt are hard-guarded; DGM explicitly leaves ``archive maintenance and parent selection \ldots\ fixed and not modifiable by the DGM'' (\S3); HyperAgents permits inner-routine edits but its outer selection and evaluation loop ``cannot be altered'' (\S7). The fraction reflects that modification of the solver (level 1) and of the modification logic itself (level 2, since that logic lives inside the source being edited) are both fully realized, whereas every level above changes only on the editable fraction of the machinery, never the frozen driver, and so earns at most half-credit.

\paragraph{Meta-scaffolding.}
Meta-Harness~\citep{metaharness2026} optimizes the code infrastructure around an LLM by exposing the full history of prior candidates' source, scores, and traces to a coding agent ($\sim$10M tokens of diagnostic context). PromptBreeder~\citep{promptbreeder2024} evolves both task-prompts and mutation-prompts (two-level self-reference in prompt space), and GEPA~\citep{agrawal2026gepa} shows that reflective prompt evolution alone can outperform RL fine-tuning, while LWE~\citep{jwa-etal-2026-becoming}
applies a reflective loop to the evaluator itself, improving
an LLM judge label-free as it evaluates. These keep the driver outside the editable surface (Meta-Harness), realize at most two meta-steps (PromptBreeder), or operate in prompt space rather than code space.

\paragraph{What \metan{} adds.}
The structural reframing yields four advantages: (i)~layered hierarchy rather than collapsed roles; (ii)~$\prod_{d=2}^{n} k_d$ compositional coverage rather than $\sum_{d=2}^{n} k_d$, since each $\Omega$-layer reshapes the execution of all layers below; (iii)~emergent depth via convergence rather than hand-set $T$ or open-ended archive growth; (iv)~a single $\Omega$ prompt template across eight heterogeneous benchmarks (orchestration hyperparameters tuned per benchmark) rather than per-benchmark prompt or operator search. Empirically, realized depths spanning 2 to 6 with emergent role specialization (generic primitives at depth~2 propagating cross-task, task-aware routing at depth~3, deeper-layer specialization and rollback) provide, to our knowledge, the first demonstration that meta-depth $>\!2$ produces structurally distinct levels rather than redundant ones.

% \begin{table}[ht]
% \caption{\textbf{Realized meta-depth across self-improving agent frameworks.} Depth counts edit layers above the task solver that are themselves modifiable at run time. See \S\ref{app:related_work_full} for which layer each system holds fixed.}
% \label{tab:comparison}
% \centering
% \resizebox{\columnwidth}{!}{%
% \begin{tabular}{lc}
% \toprule
% \textbf{System} & \textbf{Depth} \\
% \midrule
% Self-Refine~\citep{selfrefine2023}, Reflexion~\citep{reflexion2023} & 1 \\
% FunSearch~\citep{funsearch2024}, AlphaEvolve~\citep{alphaevolve2025}, OpenEvolve~\citep{openevolve} & 1 \\
% ADAS~\citep{adas2025}, AFlow~\citep{aflow2025}, Meta-Harness~\citep{metaharness2026} & 1 \\
% GEPA~\citep{agrawal2026gepa} & 1 \\
% STOP~\citep{stop2024} & 2 \\
% PromptBreeder~\citep{promptbreeder2024} & 2 \\
% EvoX~\citep{liu2026evox} & 2 \\
% G\"odel Agent~\citep{godelAgent2024}, DGM~\citep{darwinGodel2025} & 2 \\
% HyperAgents~\citep{hyperagents2026} & 2 \\
% \midrule
% \textbf{\metan{}} & \textbf{$n$ (2 to 6 observed)} \\
% \bottomrule
% \end{tabular}%
% }
% \end{table}

%==============================================================================
\section{Experimental Setup Details}
\label{app:setup_details}
%==============================================================================

\begin{table}[ht]
\centering
\setlength{\tabcolsep}{4pt}
\caption{\textbf{Per-benchmark evolutionary hyperparameters} on the two backbones. The $\Omega$ prompt template, output schema, $\epsilon=0.02$, and $\alpha=0.3$ are fixed across all rows. Max-depth is 10 except TB2 (4). ``off'' means early stopping is disabled and the run goes to max-iter. The GPT-5.2 column tightens max-iter on CO-Bench (8 to 6) and S2D (20 to 10) to bound per-seed cost, and leaves AE Math and SR at 6; TB2 sweeps category by category. ARC-AGI-2 runs under GPT-5.2 only (seeds 42/43/44; ${\sim}\$632$ per \metan{} seed).}
\label{tab:hparams}
\begin{tabular}{lcccccc}
\toprule
& & & & \multicolumn{2}{c}{\textbf{Max-iter}} & \\
\cmidrule(lr){5-6}
\textbf{Benchmark} & $B$ & $K$ & \textbf{Pat.} & \textbf{Gemma} & \textbf{GPT-5.2} & \textbf{Test split?} \\
\midrule
CO-Bench & 2 & 2 & 4 & 8 & 6 & yes \\
Symptom2Disease & 1 & 3 & off & 20 & 10 & yes \\
LawBench & 1 & 3 & off & 20 & n/r & yes \\
TerminalBench~2.0 & 1 & 2 & 3 to 4 & 6 to 10 & 13 cat-by-cat & no \\
AlphaEvolve Math & 2 & 2 & 10 & 6 & 6 & no \\
Symbolic Regression & 2 & 2 & 10 & 6 & 6 & no \\
AlgoTune & 2 & 2 & 10 & 6 & n/r & no \\
ARC-AGI-2 & 2 & 2 & 4 & n/r & 8 & yes \\
\bottomrule
\end{tabular}
\end{table}

\paragraph{Benchmarks.}
(i)~\textbf{CO-Bench}~\citep{cobench2026}: 36 NP-hard combinatorial optimization problems, Python solver, scored $[0,1]$.
(ii)~\textbf{Symptom2Disease} (S2D): 22-class medical diagnosis, accuracy metric; Python solver with an ambient \texttt{llm()} helper.
(iii)~\textbf{LawBench charge prediction}: multi-label Chinese legal classification, F1 metric; same Python+\texttt{llm()} format.
(iv)~\textbf{TerminalBench~2.0} (TB2): 89 terminal agent tasks across 13 categories in Docker-sandboxed environments, bash solver; $\Omega$ emits both bash functions and Python helper scripts.
(v to vii)~\textbf{OpenEvolve benchmarks}: AlphaEvolve Math (mathematical discovery; problem count varies by setup as reconciled below), Symbolic Regression (SR; equation discovery from data, 4 domains), AlgoTune (algorithm speedup); all integrated via a shared adapter with auto-discovery of problem directories and subprocess-isolated evaluation.
(viii)~\textbf{ARC-AGI-2}: 120 abstract-reasoning grid puzzles in which the solver must infer a transformation rule from a small number of input/output examples, scored pass@2; run under GPT-5.2 only.

\paragraph{AlphaEvolve Math problem-count reconciliation.}
The AE Math task pool was iteratively expanded as new problems were ported into the OpenEvolve adapter. The Gemma single-shot run was evaluated on the 15 problems available at run time; the Gemma agentic and the three-seed GPT-5.2 \metan{} and G\"odel Agent runs were evaluated on 16 problems; the OpenEvolve GPT-5.2 run was evaluated on a 10-problem subset because its diff-based mutation requires per-problem starter solvers and three of the newer problems lacked one. Aggregate AE Math numbers use the largest set available to each system.

\paragraph{Solver and orchestrator hyperparameters.}
Base solver temperature 0.3; $\Omega$ temperature 0.7 (linear) or cycling $\{0.5, 0.7, 0.9\}$ (evolutionary). \emph{Linear:} $\epsilon = 0.02$ (minimum mean-score gain for an injection to count as progress), $D = 10$ (max-depth cap), 10\,s evaluation timeout. \emph{Evolutionary:} the $\Omega$ prompt template, output schema, $\epsilon = 0.02$, and $\alpha = 0.3$ (exploration bonus that down-weights already-extended chains in parent sampling) are fixed across benchmarks; per-benchmark beam width $B$, branching $K$, patience, and max-iter appear in Table~\ref{tab:hparams}. Self-debug allows \texttt{max\_retries}$=2$ with a benchmark-dependent \texttt{retry\_threshold} ($0.1$ on CO-Bench and classification, $0.3$ on AlphaEvolve Math and AlgoTune, $0.5$ on Symbolic Regression and TB2). Single-shot permits up to two executor-level retries; agentic permits up to 8 observe-then-act turns. Both variants share the same evolutionary hyperparameters.

\paragraph{Wall-clock cost.}
A full evolutionary run terminates by patience or archive saturation; neither is bounded a priori. On CO-Bench (36 tasks) a single-shot run takes ${\sim}3$\,h on a single machine driving the OpenRouter API, dominated by Layer-1 evaluations rather than $\Omega$ calls, and the agentic variant takes ${\sim}10$\,h (up to 8 LLM calls per task). TB2 wall-clock is dominated by Docker start-up; the 89-task agentic run took ${\sim}28$\,h. Spend scales with the per-row token budgets in Table~\ref{tab:variants_meta}.

\begin{table}[ht]
\caption{\textbf{Per-row depth, archive size, and token spend} for the seven Gemma benchmark families, including the four panels of Figure~\ref{fig:variants} (s42 production runs).}
\label{tab:variants_meta}
\centering
\setlength{\tabcolsep}{5pt}
\begin{tabular}{llccr}
\toprule
\textbf{Benchmark} & \textbf{Mode} & \textbf{Depth} & \textbf{Cands} & \textbf{Tokens} \\
\midrule
CO-Bench (36 tasks)      & Single-shot & 3 & 33 & 5.3M \\
                         & Agentic     & 3 & 29 & 36.1M \\
\cmidrule(lr){1-5}
Symptom2Disease          & Single-shot & 5 & 61 & 422K \\
                         & Agentic     & 3 & 61 & 1.7M \\
\cmidrule(lr){1-5}
LawBench (Chinese)       & Single-shot & 4 & 61 & 761K \\
                         & Agentic     & 4 & 61 & 2.3M \\
\cmidrule(lr){1-5}
TerminalBench~2.0 (89)   & Single-shot & 4 & 92  & 3.7M \\
                         & Agentic     & 3 & 115 & 35.6M \\
\cmidrule(lr){1-5}
AlphaEvolve Math (15/16) & Single-shot & 3 & 25 & 2.9M \\
                         & Agentic     & 4 & 24 & 25.7M \\
\cmidrule(lr){1-5}
Symbolic Regression (4)  & Single-shot & 3 to 6 & 97  & 12.1M \\
                         & Agentic     & 3 to 4 & 100 & 21.7M \\
\cmidrule(lr){1-5}
AlgoTune (8)             & Single-shot & 3 & 25 & 1.1M \\
                         & Agentic     & 4 & 24 & 6.9M \\
\bottomrule
\end{tabular}
\end{table}

\paragraph{Baseline configurations (Gemma).}
\textbf{G\"odel Agent} (GA): 20 iterations, max\_val\,=\,50, n\_few\_shot\,=\,10, solve\_timeout\,=\,300\,s; same base model as \metan{}. We run GA on CO-Bench, Symptom2Disease, LawBench, AlphaEvolve Math, and AlgoTune. TB2 is excluded because running GA there would require substantial changes to its harness; SR baselines were run under GPT-5.2 only. All reported CO-Bench GA numbers use the corrected per-task configuration; the published single-solver interface and the budget-parity study are documented in Appendix~\ref{app:godel_details}.
\textbf{OpenEvolve} (OE): three islands, MAP-Elites 10$\times$10 feature grid, OpenEvolve's default temperature schedule; same base model. CO-Bench runs 36 \emph{independent} per-task evolutions with 10 iterations each and diff-based code edits; Symptom2Disease and LawBench run one full-rewrite prompt evolution per task for 20 iterations. The compute-parity runs hard-cap \metan{}'s token spend to each benchmark's measured OE budget (Appendix~\ref{app:oe_budget_controls}).

\paragraph{Baseline configurations (GPT-5.2).}
We re-ran both GA and OE with the Gemma hyperparameters on CO-Bench, Symptom2Disease, AlphaEvolve Math, SR, and ARC-AGI-2 (seeds 42/43/44), against the \metan{} cells of Table~\ref{tab:vs_baselines_gpt52}. The only deviations are the cost-bounded iteration caps in the GPT-5.2 column of Table~\ref{tab:hparams}. LawBench and AlgoTune were not run under GPT-5.2.

\paragraph{Depth-1 \metan{} ablation (GPT-5.2).}
We collapse the archive to a single layer with no inter-layer context, no code-library injection, and no recursion; $n_\text{runs}=4$ per seed at temperature $1.0$, seeds 42/43/44, ${\sim}870$K tokens per seed. No held-out test pass was run, so the depth-1 anchors in Table~\ref{tab:ablations} are archive-best validation scores (per-task best over the four runs, averaged across tasks), matching the estimator used for the full-stack anchors in that table.

\subsection{OpenEvolve budget controls}
\label{app:oe_budget_controls}

The architectural difference between \metan{} and OE is cross-task transfer. OE's per-artifact loop cannot share a discovered pattern (e.g., ``avoid scipy'') across tasks within a single run, whereas $\Omega$'s code-library injections are derived from cross-task failure patterns. On CO-Bench the two systems already run at near-identical token budget. On the prompt benchmarks, where full-budget \metan{} spends 10 to $20\times$ more tokens than OE, we hard-cap \metan{}'s total LLM-token spend to each benchmark's OE budget, on average 485K tokens (Gemma). \metan{} still leads at parity, S2D $0.732 \pm 0.023$ vs.\ $0.718 \pm 0.022$ and LawBench $0.784 \pm 0.013$ vs.\ $0.745 \pm 0.034$, so the full-budget gap contains a compute component but does not reduce to one. \metan{} is also the more \emph{sample-efficient} searcher; on CO-Bench (Gemma) it outperforms OE using ${\sim}13\times$ fewer candidate evaluations (29 vs.\ 378), a structural consequence of running one grouped search across all tasks where OE runs 36 independent per-task evolutions.

%==============================================================================
\section{Design Decisions and Rationale}
\label{app:design_rationale}
%==============================================================================

\paragraph{Same model at all layers.}
We use the same LLM at the base solver and at every $\Omega$ invocation (Gemma~4 31B-IT for the headline runs; GPT-5.2 for the cross-model study in §\ref{sec:experiments}). This isolates the contribution of \emph{recursion} from any contribution of \emph{model heterogeneity}; a depth-$d$ improvement cannot be attributed to a stronger improver. Heterogeneous mixtures (e.g.\ a stronger model at $\Omega$) are an obvious next step (§\ref{sec:conclusion}) but would confound the ablation.

\paragraph{Wrapper, not patching.}
$\Omega$'s emitted code is injected via a non-invasive \texttt{MetaLayer} wrapper; it cannot mutate the inner solver's state, the task object, or earlier layers' libraries except by name override. This rules out a class of bugs in which a self-modifying agent corrupts its own scaffolding (a failure mode reported in DGM~\S3 and the impetus for G\"odel Agent's hard-guarded action API).

\paragraph{Safety and sandboxing.}
\metan{} validates all $\Omega$-generated code in two phases. \emph{Static analysis}: (1) syntax check via \texttt{ast.parse}, (2) AST walk to detect dangerous imports (os, subprocess, socket, sys, etc.) and blocked attributes, (3) code length limit (10{,}000 chars). \emph{Smoke testing}: code library functions are exec-ed in an isolated namespace to verify they define a callable with the expected name; functions that crash at definition time (e.g., due to missing imports) are skipped with a warning rather than prepended to the solver's script.

\paragraph{3:1 failure-biased trace ratio.}
Pilot logs showed that $\Omega$ at depths 2 and 3 spends most useful inference on diagnosing failure patterns; success traces serve only as positive exemplars. We tuned \texttt{failure\_ratio} on the 10-task CO-Bench pilot; 0.5 dilutes failure signal, 1.0 strips out positive exemplars and produces overfit constraints. 0.75 is the smallest ratio that preserved the positive-exemplar role.

\paragraph{Convergence threshold $\epsilon = 0.02$.}
Below this threshold the gain is within single-seed run-to-run variance for our base model on CO-Bench, so further deepening is not justified. Evolutionary mode uses the same $\epsilon$ to define a non-improving iteration; patience triggers after $P$ consecutive such iterations.

\paragraph{Beam settings.}
For benchmarks where parents diverge meaningfully (CO-Bench, AlphaEvolve Math, AlgoTune, SR), we set $B=K=2$ (the smallest beam admitting cross-pollination). Classification (S2D, LawBench) uses $B=1, K=3$ and TB2 uses $B=1, K=2$, since prompt-only and sandbox-bound benchmarks gain less from a wider parent beam. Larger beams scale token cost linearly without changing the qualitative finding that the archive opens per-task headroom over any single chain.

\paragraph{Practical depth.}
Best-performing chains span depth~3 to 6 (plateau at 3 to 4 on most benchmarks; depth~6 on matsci symbolic regression). Optimal depth tracks failure-mode diversity. Two ceilings remain untested, the model's reasoning capacity at higher meta-levels and context-window saturation by accumulated layer code.

\paragraph{Value of the evolutionary archive.}
The archive decouples depth from quality (a depth-2 candidate can outperform a depth-3) and enables cross-candidate strategy transfer; together these open per-task headroom beyond any single chain.

%==============================================================================
\section{Extended Methodology}
\label{app:methodology}
%==============================================================================

\subsection{Evolutionary Orchestrator}
\label{app:orchestrator}

\begin{algorithm}[H]
\caption{Evolutionary Meta-Recursion}
\label{alg:evolutionary}
\resizebox{\linewidth}{!}{%
\begin{minipage}{1.2\linewidth}
\begin{algorithmic}[1]
\Require Tasks $\mathcal{T}$ with $N = |\mathcal{T}|$; base solver $S_1$; max depth $D$; beam $B$, children $K$; patience $P$; exploration $\alpha$. Chain mean $\bar S(c) = \tfrac{1}{N}\sum_t \mathrm{score}(t,c)$; archive $\mathcal{A}$ of (chain, mean) pairs; archive-best $\bar S^\star = \tfrac{1}{N}\sum_t \max_{c \in \mathcal{A}} \mathrm{score}(t,c)$; non-improving counter $p$.
  \State $\mathcal{A} \gets \{(S_1,\,\bar S_1)\}$;\; $\bar S^\star \gets \bar S_1$;\; $p \gets 0$
  \While{$p < P$}
    \State Sample $B$ parents from $\{c \in \mathcal{A} : \text{depth}(c) < D\}$ with $w(c) \propto \bar S(c) + \alpha/(1{+}\text{children}(c))$
    \For{each parent $c_p$, $k = 1, \dots, K$}
      \State $C \gets \Omega(c_p)$ \Comment{$T$ cycled through $\{0.5,0.7,0.9\}$; rival traces appended for tasks where $c_p$ underperforms}
      \If{$C = \emptyset$} \textbf{continue} \EndIf
      \State $\mathcal{A} \gets \mathcal{A} \cup \{(\textsc{MetaLayer}(C,\, c_p),\,\bar S_{\text{new}})\}$
    \EndFor
    \State $\bar S^\star_{\text{cur}} \gets \tfrac{1}{N}\sum_t \max_{c \in \mathcal{A}} \mathrm{score}(t, c)$
    \State $p \gets 0$ if $\bar S^\star_{\text{cur}} > \bar S^\star$ else $p{+}1$;\; $\bar S^\star \gets \max(\bar S^\star, \bar S^\star_{\text{cur}})$
  \EndWhile
  \State \Return $\arg\max_{c \in \mathcal{A}} \bar S(c)$;\; archive-best $\bar S^\star$
\end{algorithmic}
\end{minipage}%
}
\end{algorithm}

\subsection{Trace Sampling and Depth-Aware Payload}
\label{app:trace_payload}

$\Omega$'s LLM call template (system prompt, output schema, parsing) is identical at every depth; only the trace payload it ingests adapts to the available signal. Each call sees at most 20 traces drawn at a 3:1 failure-to-success ratio (65\% of the context window is reserved for traces, 35\% for the prior code stack; oldest layers drop first on overflow).

\paragraph{Depth $\leq 2$: raw per-task traces.}
At depth 2 the payload is per-task formatted traces plus a baseline-comparison flag marking improved or regressed tasks. Each trace contains the generated script, stdout/stderr (truncated to 500 chars), exit code, an error summary, and the evaluator feedback string.

\paragraph{Depth $\geq 3$: structured summary.}
At depth $\geq 3$ raw traces are replaced by a structured summary: (1) a task-category performance breakdown, (2) failure-pattern distribution, (3) an effectiveness analysis of the previous layer measured against depth $d{-}2$ to isolate its contribution, and (4) three representative traces. In evolutionary mode, regressed tasks are additionally annotated with the archive's per-task best, giving $\Omega$ an explicit improvement ceiling.

The depth-aware formatter is therefore a content adapter above $\Omega$, not a separate driver; $\Omega$'s prompt template and output parser do not change as depth grows.

\subsection{Pattern Categorization Rubric and Inter-Rater Agreement}
\label{app:patterns_rubric}

Each of the 596 $\Omega$ emissions across the CO-Bench, S2D, LawBench, AlphaEvolve Math, and SR production archives (119 at depth 2, 203 at depth 3, 180 at depth 4, 86 at depth 5, 8 at depth 6) is classified into one or more of seven categories: (i)~environment constraints (e.g., library exclusions); (ii)~tactical primitives (callable library functions such as \texttt{local\_search()}); (iii)~task-category routing (\texttt{if/elif} over task type or topic); (iv)~specialized libraries (task-typed helpers such as \texttt{geometry\_utils()}); (v)~prompt-engineering edits (role, language, CoT scaffolding); (vi)~rollback or override (rationale text invoking ``regress'', ``rollback'', ``over-constrain'', ``abandon''); (vii)~compositional architectures (multi-function pipelines).

\begin{figure}[ht]
\centering
\begin{subfigure}[t]{0.468\linewidth}
    \centering
    \includegraphics[width=\linewidth]{figures/emergent_roles_code_substrates.pdf}
    \caption{\textbf{Code substrates} (CO-Bench, AlphaEvolve Math, SR). Denominators are $n = 68, 125, 72, 23$ emissions at depths 2 to 5. Tactical primitives peak at depth 3 and then give way to specialized libraries; prompt engineering is near zero throughout, since there is no prompt to rewrite.}
    \label{fig:emergent_roles:code}
\end{subfigure}\hfill
\begin{subfigure}[t]{0.513\linewidth}
    \centering
    \includegraphics[width=\linewidth]{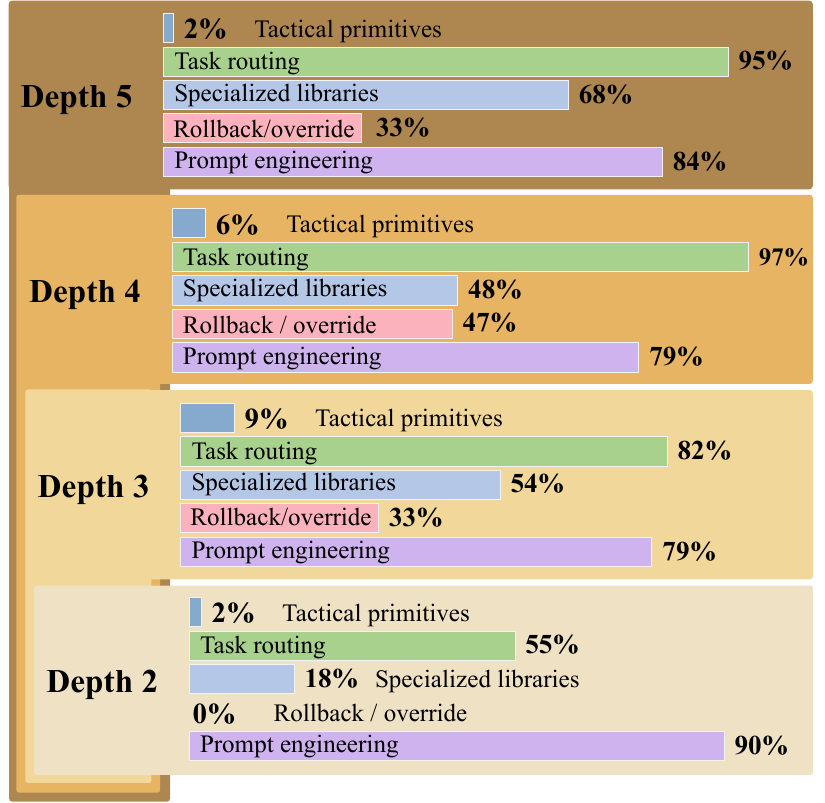}
    \caption{\textbf{Prompt substrates} (Symptom2Disease, LawBench). Denominators are $n = 51, 78, 108, 63$. Prompt engineering saturates at every depth and tactical primitives never exceed 9\%, the mirror image of (a). The action surface is text, so that is where $\Omega$ works.}
    \label{fig:emergent_roles:prompt}
\end{subfigure}
\caption{\textbf{Emergent layer roles by depth, split by solver substrate.} Each bar is the fraction of that depth's $n$ emissions ($n$ per depth in the subcaptions) that \emph{both} raters, GPT-5.2 and Kimi-K2.6, assign to the category, over the 588 emissions at depths 2 to 5. The rubric is multi-label, so bars within a depth sum past 100\% and should be compared across depths rather than against one another. Emissions the raters split on are discarded, making every bar a lower bound. Two negligible categories are omitted; rubric and per-category $\kappa$ in Appendix~\ref{app:patterns_rubric}.}
\label{fig:emergent_roles:full}
\end{figure}

Two independently prompted LLM raters, GPT-5.2 and Kimi-K2.6, annotated all 596 emissions under this rubric. Figure~\ref{fig:emergent_roles:full} reports the fraction of emissions at each depth that \emph{both} raters mark as matching a category; an emission may match several categories, so fractions within a depth do not sum to 100\%. The figure omits the two categories with negligible mass. Environment constraints are a shallow-code phenomenon that decays with depth (13\%, 11\%, 3\%, 9\% of code-substrate emissions at depths 2 to 5, i.e.\ 9, 14, 2 and 2 emissions), consistent with the depth-2 dependency workaround in the worked example of \S\ref{sec:intro}. Compositional architectures stay below 14\% everywhere, and their apparent rise on deep code layers rests on 3 emissions out of 23, so we do not read a trend into it. Table~\ref{tab:kappa} gives the per-category Cohen's $\kappa$. Agreement is near-perfect where the cue is close to syntactic (task routing, prompt engineering) and weakest on the two most abstract roles, where GPT-5.2 labels more liberally than Kimi. Both headline patterns of \S\ref{sec:case_study}, the depth emergence of the corrective role and the code-vs-text split, reproduce under either rater alone.

The full substrate comparison complements the code-focused figure in Section~\ref{sec:case_study}. Rollback is absent at depth 2 in both substrates under both raters and appears at depth 3 (55\% on code, 33\% on prompt). Code benchmarks show a shift from tactical primitives toward specialized libraries, while prompt benchmarks remain dominated by prompt engineering; role progression therefore depends on the solver's action space.

\begin{table}[ht]
\caption{\textbf{Inter-rater agreement per rubric category} (Cohen's $\kappa$, GPT-5.2 vs.\ Kimi-K2.6, all 596 emissions).}
\label{tab:kappa}
\centering
\small
\begin{tabular}{lc}
\toprule
\textbf{Category} & $\kappa$ \\
\midrule
Task routing & 0.96 \\
Prompt engineering & 0.85 \\
Rollback / override & 0.60 \\
Specialized libraries & 0.57 \\
Environment constraints & 0.45 \\
Tactical primitives & 0.37 \\
Compositional architectures & 0.30 \\
\midrule
Mean over the 7 categories & 0.59 \\
\bottomrule
\end{tabular}
\end{table}

\subsection{Depth-wise task analysis and applicability}
\label{app:depth_analysis}

This analysis expands the summary in Section~\ref{sec:case_study}, with task-level examples of transfer, interference, and useful depth, followed by the conditions under which the framework helps.

\paragraph{Depth 2: generic primitives that propagate across tasks.}
On CO-Bench, depth-2 emissions are wholly generic; both raters mark all 12 of its d2 emissions as tactical primitives (\texttt{local\_search}, \texttt{simulated\_annealing}, \texttt{normalize\_label}, \texttt{safe\_log}). The lower 28\% in Figure~\ref{fig:emergent_roles:code} is dilution from the other two code archives, where depth 2 more often routes. A majority of CO-Bench winners ultimately call the primitives, with 22 of 36 winning solvers invoking at least one $\Omega$-emitted library function. \texttt{simulated\_annealing}, first emitted at depth~2 in a single candidate, propagates to 15 of 36 winners across four problem families. Every chain improves $d1\!\to\!d2$ (mean lift $+0.113$); depth 2 alone fixes a mean of 3.7 previously-failing tasks per chain.

\paragraph{Depth 3: specialization and layer interference.}
Depth 3 leans into specialized prescriptions for particular task families, and this is where the layer interference anticipated in \S\ref{sec:method:recursion} first appears, since a prescription tuned to one family can override guidance that was working elsewhere. On code substrates, 68\% of d3 emissions ship specialized libraries, and task-category \texttt{if/elif} routing intensifies with depth rather than debuting here (Figure~\ref{fig:emergent_roles:code}); the aggregate $d2\!\to\!d3$ mean lift is only $-0.006$, but per-task variation is large, with 41\% of (chain, task) pairs strictly regressing at depth 3 and 18\% dropping substantially (d3 ${<}\,0.8\cdot$d2). On \texttt{multi\_demand\_multidim\_knapsack}, depth 2's ``greedy + \texttt{local\_search}'' lifted $0.0\!\to\!0.759$ while depth 3's ``\texttt{simulated\_annealing} with group-transfer move'' regressed to $0.230$; the kind of failure that motivates the rollback behaviour at depth 4.

\paragraph{Deeper layers (4 to 6): corrections and refinements.}
At depth $\geq 4$ the corrective and refining roles persist. On code substrates, rollback intent holds near half of emissions at both depths while specialized libraries keep rising, and on prompt substrates, rollback peaks at depth 4. Deeper layers therefore correct and refine prior layers rather than redesign the solver. Mean score per depth peaks at depth 3 ($0.726$) and falls to $0.602$ by depth 6, but a per-depth mean is not the quantity an archive optimizes. \textbf{Deep layers keep producing per-task wins that no shallower chain reaches.} Depth-${\geq}4$ candidates win 31\% of CO-Bench tasks, 80\% of SR-matsci, and 70\% of SR-phys-osc. \texttt{constrained\_guillotine\_cutting} first opens at depth 5 (all ancestors score 0); three matsci tasks first open at depth 6. The archive thus preserves both shallower fallbacks (for d3 over-specification) and rare deep-layer specialists.

\paragraph{The framework is universal; the role progression adapts.}
A single $\Omega$ template runs across all eight benchmarks, but the emergent role progression adapts to each benchmark's failure landscape. It is most pronounced where failure modes are diverse (CO-Bench, SR-matsci, SR-phys-osc); prompt-only benchmarks (S2D, LawBench) progress within prompt-engineering space at every depth, since the action surface is just text.

\paragraph{Where the framework pays off.}
How much \metan{} gains tracks how many distinct ways a benchmark can fail. CO-Bench's 36 NP-hard shapes and SR-matsci's 25 distinct laws each present many orthogonal failure modes, so the cross-task code library and per-depth specialization compound into the largest margins we see; TB2's 13 sandbox categories and AlphaEvolve Math's mixed numerical-combinatorial tasks behave the same way. ARC-AGI-2 is the extreme case, where the baselines barely move and recursion produces the only above-floor entries. S2D sits at the other end, since its small action surface (one prompt, 22 labels) holds \metan{}'s lead over OE to within seed noise on both backbones. The dev-to-test gradient shows the same ordering. CO-Bench (dev $0.886$, test $0.870$) transfers archive structure to held-out tasks; S2D (dev $0.887$, test $0.725$) saturates as the prompt-rewrite contract collapses onto a single held-out distribution; ARC-AGI-2 preserves the categorical above-floor signal even as the absolute number falls.

\paragraph{Where it does not.}
A second condition is the room between the seed and the ceiling, and two benchmarks sit at its limit. AlgoTune is the over-constraint case, where a pre-optimized kernel contract leaves little for $\Omega$ to add and its extra context tightens the code rather than improving it. Most of the slowdown comes from the two FFT kernels rather than being spread across the benchmark (Appendix~\ref{app:algotune}), so the diagnosis can be checked task by task instead of resting on a single aggregate number. On SWE-Bench ($B{=}K{=}2$, three iterations) the seed is already strong enough that the archive's best is the generation-0 candidate and $\Omega$ never activates. $\Omega$'s value therefore scales with both quantities, the headroom above the seed and the diversity of failure modes the archive can index.

\subsection{Convergence behavior}
\label{app:convergence}

\begin{figure}[ht]
\centering
\includegraphics[width=0.9\textwidth]{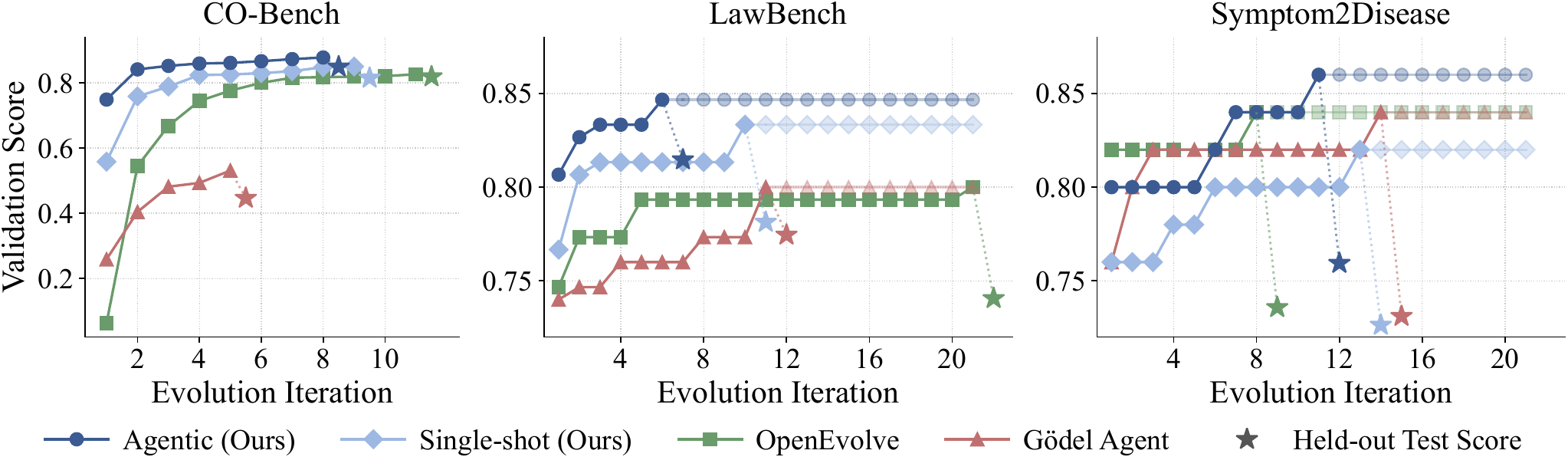}
\caption{\textbf{Search progress on CO-Bench, LawBench, and Symptom2Disease} (Gemma~4 31B-IT, single-seed runs). Lines are best-so-far validation over evolution iterations; stars ($\star$) are held-out test scores plotted at a fixed offset; faded markers hold the last value after early stopping.}
\label{fig:convergence_3way}
\end{figure}

Figure~\ref{fig:convergence_3way} shows \metan{} agentic reaching higher scores in fewer evolution iterations than both baselines on all three benchmarks, while the single-shot variant trails both on S2D. OE climbs but settles below, and GA rises quickly before flattening near $0.53$ on CO-Bench, well short of \metan{}. Raising GA's budget does not close that gap (\S\ref{sec:parity}).

%==============================================================================
\section{Depth 2 Injected Code (Full)}
\label{app:depth2_code}
%==============================================================================

The following is the complete pre-process code generated by $\Omega$ at depth 2 during the pilot experiment:

{\small
\begin{verbatim}
# Address "No module named 'scipy'" and timeout patterns.
additional_context = """
### IMPORTANT ENVIRONMENT RESTRICTIONS:
1. EXTERNAL LIBRARIES: Do NOT use `scipy` or any
   third-party libraries except for `numpy`. If you need
   linear programming, assignment algorithms (like
   Hungarian), or optimization solvers, you MUST implement
   them from scratch using pure Python or `numpy`.
2. PERFORMANCE: Avoid nested loops over large coordinate
   ranges. Use coordinate compression, event-based
   processing, or greedy heuristics with sparse data
   structures to avoid Timeouts.
3. VALIDATION: Ensure that all output indices strictly
   match the indexing requirements specified in the task
   description.
"""
\end{verbatim}
}

This single piece of code, generated automatically by $\Omega$ from failure trace analysis, improved mean score from 0.439 to 0.513 (+16.9\% relative) by preventing the most common failure mode across 4 of 10 tasks.

%==============================================================================
\section{Consolidation Guard: Study and Control}
\label{app:consolidation}
%==============================================================================

Deeper layers can regress individual tasks (§\ref{sec:method:recursion}); a depth-3 router that misfires on a task family overrides depth-2 guidance that was working. The consolidation mode of §\ref{sec:method:orchestration} guards against this by construction. Each candidate targets a single \emph{focus task} and inherits the archive's frozen best traces for every other task, so the per-task-best trajectory can never decrease.

\paragraph{Study.}
On an 8-task CO-Bench band (3 seeds), consolidation mode lifted the per-task-best mean from 0.502 to $0.71 \pm 0.02$ with \emph{zero} per-task regressions on 8 of 8 tasks. A compute-matched control, best-of-4 seed resampling under the standard orchestrator, reaches $0.61 \pm 0.03$; the $+0.10$ lift over the control carries a 95\% confidence interval of $[+0.04, +0.16]$. An earlier single-seed pilot (6-task subset, 8 generations, locally served quantized Gemma backbone) showed the same zero-regression behaviour ($0.502 \to 0.685$ vs.\ control $0.674$).

\paragraph{What we claim.}
On this band consolidation also beats the compute-matched control on mean lift ($+0.10$, confidence interval excluding zero), but we rest the claim on the \emph{shape} of the trajectory; the unguarded orchestrator produces per-task regressions at a substantial pair-level rate when depth-3 routing misfires, whereas consolidation is monotone by construction and can never regress an already-solved task. The guard therefore matters most in deployments where such a regression is costlier than a slower mean climb.

%==============================================================================
\section{Symbolic Regression Per-Domain Breakdown}
\label{app:sr_per_domain}
%==============================================================================

Table~\ref{tab:sr_per_domain} expands the SR entries of Tables~\ref{tab:vs_baselines} and~\ref{tab:vs_baselines_gpt52} into the four per-domain evolutions. On the Gemma production runs, single-shot wins archive-best on \texttt{chem\_react} and \texttt{phys\_osc}, whose stronger seeds leave limited headroom; agentic wins \texttt{bio\_pop\_growth}, \texttt{matsci}, and the family mean. Under GPT-5.2 (Table~\ref{tab:sr_per_domain_gpt52}) \metan{} leads every domain against both baselines. G\"odel Agent's large stdevs (e.g., \texttt{chem\_react} $2.92 \pm 3.84$) come from seed-level collapses in which its single self-modification regresses the solver.

\begin{table}[ht]
\caption{\textbf{Symbolic Regression per-domain results (Gemma, s42 production runs),} backing the Symbolic Regression row of Table~\ref{tab:variants_meta}. Domains: \texttt{bio\_pop\_growth} (biological population dynamics), \texttt{chem\_react} (chemical reaction kinetics), \texttt{matsci} (materials science properties), \texttt{phys\_osc} (physical oscillator systems). Bottom block gives the family mean (Tokens summed across the four runs; other columns averaged). Bold = higher archive-best per domain.}
\label{tab:sr_per_domain}
\centering
\small
\setlength{\tabcolsep}{4pt}
\resizebox{\textwidth}{!}{%
\begin{tabular}{llcccccr}
\toprule
\textbf{Domain} & \textbf{Mode} & \textbf{Seed} & \textbf{Linear-best} & \textbf{Archive-best} & \textbf{Depth} & \textbf{Cands} & \textbf{Tokens} \\
\midrule
bio\_pop\_growth (24 inst.) & Single-shot & 0.477 & 2.751 & 3.782 & 3 & 25 & 2.6M \\
                            & Agentic     & 2.355 & 4.406 & \textbf{4.977} & 3 & 25 & 5.2M \\
\cmidrule(lr){1-8}
chem\_react (36 inst.)      & Single-shot & 5.458 & 6.580 & \textbf{7.599} & 3 & 25 & 2.6M \\
                            & Agentic     & 5.587 & 6.440 & 7.244 & 3 & 25 & 2.7M \\
\cmidrule(lr){1-8}
matsci (25 inst.)           & Single-shot & 0.000 & 1.273 & 2.130 & 6 & 23 & 3.9M \\
                            & Agentic     & 1.477 & 3.296 & \textbf{3.863} & 3 & 25 & 6.7M \\
\cmidrule(lr){1-8}
phys\_osc (44 inst.)        & Single-shot & 2.686 & 4.191 & \textbf{4.681} & 4 & 24 & 3.0M \\
                            & Agentic     & 3.114 & 3.573 & 4.514 & 4 & 25 & 7.1M \\
\midrule
\emph{Mean across 4 domains} & Single-shot & 2.155 & 3.699 & 4.548 & n/a & n/a & 12.1M \\
                             & Agentic     & 3.133 & 4.429 & \textbf{5.149} & n/a & n/a & 21.7M \\
\bottomrule
\end{tabular}%
}
\end{table}

\begin{table}[ht]
\caption{\textbf{Symbolic Regression per-domain comparison under GPT-5.2} (agentic archive-best, mean $\pm$ stdev over seeds 42/43/44).}
\label{tab:sr_per_domain_gpt52}
\centering
\small
\setlength{\tabcolsep}{5pt}
\begin{tabular}{lccc}
\toprule
\textbf{Domain} & \textbf{\metan{}} & \textbf{OpenEvolve} & \textbf{G\"odel Agent} \\
\midrule
bio\_pop\_growth & \textbf{4.35 $\pm$ 0.30} & 2.83 $\pm$ 0.08 & 2.68 $\pm$ 0.83 \\
chem\_react      & \textbf{7.65 $\pm$ 0.07} & 6.96 $\pm$ 0.08 & 2.92 $\pm$ 3.84 \\
matsci           & \textbf{3.26 $\pm$ 0.67} & 1.37 $\pm$ 0.34 & 1.27 $\pm$ 0.67 \\
phys\_osc        & \textbf{4.85 $\pm$ 0.11} & 3.63 $\pm$ 0.15 & 2.93 $\pm$ 2.03 \\
\midrule
Mean             & \textbf{5.03 $\pm$ 0.20} & 3.70 $\pm$ 0.13 & 2.45 $\pm$ 1.63 \\
\bottomrule
\end{tabular}
\end{table}

%==============================================================================
\section{AlgoTune Per-Task Breakdown}
\label{app:algotune}
%==============================================================================

Table~\ref{tab:algotune_per_task} breaks the AlgoTune aggregate into per-task speedups for the two \metan{} variants (single seed, s42). The inversion is largest on the two FFT kernels, which alone account for 11.08 of the 17.55 total speedup lost across the seven shared tasks. Two further tasks lose a third to a half of their speedup under the agentic variant, psd\_cone\_projection (5.54 to 3.74) and affine\_transform\_2d (2.00 to 1.05), and the remaining three are near parity. In the single-shot run, the depth-2 prescriptive hints regressed \texttt{fft\_convolution} to $1.11\times$, and the depth-3 rationale diagnosed the cause, a missing memory-contiguity constraint, and repaired it to $9.72\times$. The agentic run never recovered the same kernel. This is the task-level evidence for the over-constraint diagnosis in \S\ref{sec:case_study}.

\begin{table}[ht]
\caption{\textbf{AlgoTune per-task speedups} ($\times$, s42). The seven tasks solved by both variants. The eighth task, \texttt{eigenvectors\_complex}, has a per-task log only in the agentic run ($\times1.11$), so the single-shot aggregate ($\times18.47$) averages seven tasks and the agentic aggregate ($\times14.11$) averages eight; on the matched seven the agentic mean is $\times15.96$.}
\label{tab:algotune_per_task}
\centering
\small
\setlength{\tabcolsep}{5pt}
\begin{tabular}{lcc}
\toprule
\textbf{Task} & \textbf{Single-shot} & \textbf{Agentic} \\
\midrule
convolve2d\_full\_fill     & 105.37 & 101.79 \\
fft\_convolution           & 9.72   & 1.69 \\
fft\_cmplx\_scipy\_fftpack & 4.17   & 1.12 \\
psd\_cone\_projection      & 5.54   & 3.74 \\
affine\_transform\_2d      & 2.00   & 1.05 \\
lu\_factorization          & 1.47   & 1.34 \\
polynomial\_real           & 1.03   & 1.02 \\
\bottomrule
\end{tabular}
\end{table}

%==============================================================================
\section{G\"odel Agent Baseline Details}
\label{app:godel_details}
%==============================================================================

\paragraph{Two configurations.}
GA's published harness seeds a single solver function from a generic math-QA template and asks it to cover all 36 heterogeneous CO-Bench tasks. The seed emits no per-instance \texttt{solve()}, so the run collapses to ${\sim}0.000$ on both backbones; the collapse is a property of the interface rather than of the model, and it reproduces exactly across seeds (under the published interface on GPT-5.2, the per-seed archive-best means are byte-identical, stdev $=0$). Switching one flag to per-task mode, with the same agent, model, and evaluator, gives each task its own solver lineage and recovers $0.451 \pm 0.023$ on Gemma and $0.527 \pm 0.033$ on GPT-5.2. All GA numbers in the main tables use the per-task configuration.

\paragraph{Budget parity.}
Raising the per-task GA budget $5\times$ and $10\times$ on GPT-5.2, seed 42, by increasing max-evolve from 4 to 20 and to 40 raises token spend to 5.6M and 11.4M per seed, but lifts held-out CO-Bench test only from $0.502$, that seed's default-budget score within the $0.527 \pm 0.033$ three-seed mean above, to $0.615$ and $0.628$. The per-iteration curve oscillates in the $0.4$ to $0.6$ band with no upward trend toward \metan{}'s $0.870$ at 17M tokens, so we did not extend parity further.

\paragraph{Pattern of self-edits.}
In the published-interface Gemma runs, GA invoked its self-modification action 4 times on Symptom2Disease (two successful prompt-level tweaks, two validator-rejected), 6 times on LawBench (progressively more ambitious multi-step prompt pipelines), once on AlphaEvolve Math, once on CO-Bench, and zero times across AlgoTune's iterations, where all seven per-iteration solver snapshots are byte-identical. The lone AlphaEvolve mutation kept the seed's integer-answer requirement although the task expects Python-module source, and the score regressed from $0.372$ to $0.079$. GA rewrites its monolithic solver in place and must choose to do so, whereas $\Omega$ always emits an injection, adding library functions and pre-process strings around an untouched inner solver (Appendix~\ref{app:depth2_code}).

%% file: checklist.tex
\section*{NeurIPS Paper Checklist}

\begin{enumerate}

\item {\bf Claims}
    \item[] Question: Do the main claims made in the abstract and introduction accurately reflect the paper's contributions and scope?
    \item[] Answer: \answerYes{}
    \item[] Justification: Sections~\ref{sec:intro} and~\ref{sec:method} state the recursive architecture and evolutionary orchestration contributions, which are evaluated across eight benchmark families and two backbones in Section~\ref{sec:experiments}. Tables~\ref{tab:vs_baselines},~\ref{tab:vs_baselines_gpt52}, and~\ref{tab:ablations} support the reported comparisons and ablations, with benchmark-specific exceptions and limitations discussed in Sections~\ref{sec:experiments} and~\ref{sec:conclusion}.

\item {\bf Limitations}
    \item[] Question: Does the paper discuss the limitations of the work performed by the authors?
    \item[] Answer: \answerYes{}
    \item[] Justification: Section~\ref{sec:conclusion} discusses the same-model-at-all-layers restriction, untested heterogeneous model combinations, and limits on useful depth; Appendix~\ref{app:design_rationale} discusses the untested reasoning-capacity and context-window ceilings. Sections~\ref{sec:experiments} and~\ref{sec:case_study} describe benchmarks with limited headroom, regressions, and incomplete baseline coverage, while Appendix~\ref{app:setup_details} documents computational costs and evaluation-set differences.

\item {\bf Theory assumptions and proofs}
    \item[] Question: For each theoretical result, does the paper provide the full set of assumptions and a complete (and correct) proof?
    \item[] Answer: \answerNA{}
    \item[] Justification: The paper does not present formal theorems. The compositional-coverage comparison ($\prod_d k_d$ versus $\sum_d k_d$) in Section~\ref{sec:method:recursion} is an upper bound on expressible configurations, and the information-set discussion is an architectural argument rather than a formal theoretical result.

    \item {\bf Experimental result reproducibility}
    \item[] Question: Does the paper fully disclose all the information needed to reproduce the main experimental results of the paper to the extent that it affects the main claims and/or conclusions of the paper (regardless of whether the code and data are provided or not)?
    \item[] Answer: \answerYes{}
    \item[] Justification: Section~\ref{sec:experiments} and Appendix~\ref{app:setup_details} describe the eight benchmark families, Gemma~4 31B-IT and GPT-5.2 backbones, evaluation protocols, seeds, and solver, orchestrator, and baseline configurations. Section~\ref{sec:method}, Appendix~\ref{app:methodology}, Algorithm~\ref{alg:evolutionary}, and the public repository at \url{https://github.com/minnesotanlp/meta-n} provide the method details, implementation, configurations, dependencies, and reproduction instructions.

\item {\bf Open access to data and code}
    \item[] Question: Does the paper provide open access to the data and code, with sufficient instructions to faithfully reproduce the main experimental results, as described in supplemental material?
    \item[] Answer: \answerYes{}
    \item[] Justification: The code, benchmark adapters, baseline implementations, configurations, dependencies, and reproduction instructions are available at \url{https://github.com/minnesotanlp/meta-n}. Section~\ref{sec:experiments} and Appendix~\ref{app:setup_details} identify the existing benchmarks and describe their task sets, evaluation protocols, and any system-specific coverage differences.

\item {\bf Experimental setting/details}
    \item[] Question: Does the paper specify all the training and test details (e.g., data splits, hyperparameters, how they were chosen, type of optimizer) necessary to understand the results?
    \item[] Answer: \answerYes{}
    \item[] Justification: Section~\ref{sec:experiments} specifies the two backbones, eight benchmark families, evaluation metrics, held-out versus benchmark-score reporting, and seed policy. Appendix~\ref{app:setup_details} gives per-benchmark search budgets, temperatures, retry and turn limits, baseline configurations, and the AlphaEvolve Math task-pool reconciliation; Appendix~\ref{app:design_rationale} explains the principal hyperparameter choices.

\item {\bf Experiment statistical significance}
    \item[] Question: Does the paper report error bars suitably and correctly defined or other appropriate information about the statistical significance of the experiments?
    \item[] Answer: \answerYes{}
    \item[] Justification: Section~\ref{sec:experiments}, Tables~\ref{tab:vs_baselines} and~\ref{tab:vs_baselines_gpt52}, and Figure~\ref{fig:variants} report means and standard deviations over seeds 42/43/44, with single-seed results and TB2's across-category rather than across-seed variability explicitly distinguished. Appendix~\ref{app:consolidation} additionally reports a three-seed control comparison with a 95\% confidence interval; overlapping seed ranges are acknowledged rather than presented as evidence of significance.

\item {\bf Experiments compute resources}
    \item[] Question: For each experiment, does the paper provide sufficient information on the computer resources (type of compute workers, memory, time of execution) needed to reproduce the experiments?
    \item[] Answer: \answerNo{}
    \item[] Justification: Appendix~\ref{app:setup_details} and Table~\ref{tab:variants_meta} report the API-based execution setup, representative wall-clock times, and per-benchmark token costs, with additional budget comparisons in Section~\ref{sec:parity}. The paper does not provide a complete per-experiment hardware and memory specification or a total compute accounting that includes all pilot and unsuccessful runs.
    
\item {\bf Code of ethics}
    \item[] Question: Does the research conducted in the paper conform, in every respect, with the NeurIPS Code of Ethics \url{https://neurips.cc/public/EthicsGuidelines}?
    \item[] Answer: \answerYes{}
    \item[] Justification: The research uses existing benchmark assets without recruiting human participants or collecting new private data. Prior work is credited, limitations are discussed, and code-execution safeguards are described in Appendix~\ref{app:design_rationale}.

\item {\bf Broader impacts}
    \item[] Question: Does the paper discuss both potential positive societal impacts and negative societal impacts of the work performed?
    \item[] Answer: \answerYes{}
    \item[] Justification: Automated program search can aid scientific discovery and optimization, but may produce unsafe or incorrect programs and incurs computational costs. Appendix~\ref{app:design_rationale} describes execution safeguards; benchmark performance does not establish suitability for medical or legal deployment.
    
\item {\bf Safeguards}
    \item[] Question: Does the paper describe safeguards that have been put in place for responsible release of data or models that have a high risk for misuse (e.g., pre-trained language models, image generators, or scraped datasets)?
    \item[] Answer: \answerNA{}
    \item[] Justification: The release consists of framework and experiment code, not newly trained models or scraped datasets. Code-execution safeguards are described in Appendix~\ref{app:design_rationale}.

\item {\bf Licenses for existing assets}
    \item[] Question: Are the creators or original owners of assets (e.g., code, data, models), used in the paper, properly credited and are the license and terms of use explicitly mentioned and properly respected?
    \item[] Answer: \answerNo{}
    \item[] Justification: Benchmark and baseline sources are identified in Section~\ref{sec:experiments} and Appendix~\ref{app:setup_details}. However, the paper does not enumerate license names, versions, and terms for every external code, data, and model asset.

\item {\bf New assets}
    \item[] Question: Are new assets introduced in the paper well documented and is the documentation provided alongside the assets?
    \item[] Answer: \answerYes{}
    \item[] Justification: The MIT-licensed framework, experiment configurations, dependencies, and reproduction instructions are available at \url{https://github.com/minnesotanlp/meta-n}. The architecture and experimental protocols are documented in Section~\ref{sec:method} and Appendices~\ref{app:setup_details} and~\ref{app:methodology}.

\item {\bf Crowdsourcing and research with human subjects}
    \item[] Question: For crowdsourcing experiments and research with human subjects, does the paper include the full text of instructions given to participants and screenshots, if applicable, as well as details about compensation (if any)? 
    \item[] Answer: \answerNA{}
    \item[] Justification: The research does not involve crowdsourcing or experiments with recruited human participants; the layer-role annotations are produced by LLM raters as described in Appendix~\ref{app:patterns_rubric}.

\item {\bf Institutional review board (IRB) approvals or equivalent for research with human subjects}
    \item[] Question: Does the paper describe potential risks incurred by study participants, whether such risks were disclosed to the subjects, and whether Institutional Review Board (IRB) approvals (or an equivalent approval/review based on the requirements of your country or institution) were obtained?
    \item[] Answer: \answerNA{}
    \item[] Justification: The research does not involve experiments with human participants or the collection of new participant data, so human-subject study approvals are not applicable.

\item {\bf Declaration of LLM usage}
    \item[] Question: Does the paper describe the usage of LLMs if it is an important, original, or non-standard component of the core methods in this research? Note that if the LLM is used only for writing, editing, or formatting purposes and does \emph{not} impact the core methodology, scientific rigor, or originality of the research, declaration is not required.
    %this research? 
    \item[] Answer: \answerYes{}
    \item[] Justification: Section~\ref{sec:method} describes the LLM-based base solver and fixed $\Omega$ meta-operation, while Section~\ref{sec:experiments} and Appendix~\ref{app:setup_details} specify the Gemma~4 31B-IT and GPT-5.2 backbones and inference settings. Appendix~\ref{app:patterns_rubric} also discloses the use of GPT-5.2 and Kimi-K2.6 as layer-role annotation raters.

\end{enumerate}